\documentclass[10pt, journal]{ieeeconf}
\IEEEoverridecommandlockouts

\usepackage{cite}
\usepackage{amsmath,amssymb,amsfonts}
\usepackage{algorithmic}
\usepackage{graphicx}
\usepackage{subcaption}
\usepackage{textcomp}
\usepackage{multirow}
\usepackage{xcolor}
\usepackage{color, colortbl}
\definecolor{Gray}{gray}{0.9}
\usepackage{bm}
\usepackage{booktabs}
\usepackage{tikz}
\usepackage{hyperref}
\usepackage[T1]{fontenc}

\def\BibTeX{{\rm B\kern-.05em{\sc i\kern-.025em b}\kern-.08em
    T\kern-.1667em\lower.7ex\hbox{E}\kern-.125emX}}
\usetikzlibrary{arrows,decorations.pathmorphing,backgrounds,fit,positioning,calc,shapes}

\usepackage[linesnumbered,ruled,algo2e,resetcount]{algorithm2e}

\begin{document}

\title{Comparison between Behavior Trees and Finite State Machines}

\author{\authorblockN{
Matteo Iovino$^{a}$,
Julian F\"orster$^{b}$,
Pietro Falco$^{c}$,
Jen Jen Chung$^{d}$,
Roland Siegwart$^{b}$, and
Christian Smith$^{e}$}
\thanks{
\textbf{This work has been submitted to the IEEE for possible publication. Copyright may be transferred without notice, after which this version may no longer be accessible.}
This project is financially supported by the Swedish Foundation for Strategic Research. The authors gratefully acknowledge this support.}
\thanks{$^{a}$ABB Corporate Research, Västerås, Sweden}
\thanks{$^{b}$Autonomous Systems Lab, ETH Zürich, Zürich, Switzerland}
\thanks{$^{c}$Department of Information Engineering, University of Padova, Italy}
\thanks{$^{d}$School of EECS, The University of Queensland, Australia}
\thanks{$^{e}$Division of Robotics, Perception and Learning, KTH - Royal Institute of Technology, Stockholm, Sweden}
}

\maketitle

\begin{abstract}
Behavior Trees (BTs) were first conceived in the computer games industry as a tool to model agent behavior, but they received interest also in the robotics community as an alternative policy design to Finite State Machines (FSMs). The advantages of BTs over FSMs had been highlighted in many works, but there is no thorough practical comparison of the two designs. Such a comparison is particularly relevant in the robotic industry, where FSMs have been the state-of-the-art policy representation for robot control for many years. In this work we shed light on this matter by comparing how BTs and FSMs behave when controlling a robot in a mobile manipulation task. The comparison is made in terms of reactivity, modularity, readability, and design. We propose metrics for each of these properties, being aware that while some are tangible and objective, others are more subjective and implementation dependent. The practical comparison is performed in a simulation environment with validation on a real robot. We find that although the robot's behavior during task solving is independent on the policy representation, maintaining a BT rather than an FSM becomes easier as the task increases in complexity.
\end{abstract}


\section{Introduction}

In modern industrial applications robots share their environment with humans, so they need to handle the unpredictability that may arise from unexpected outcomes of actions or different types of failures and errors. To this end, robots need to be controlled by reactive policies since fixed sequences of actions may lead to faulty behaviors.
These modern manufacturing environments require flexible policies that can adapt rapidly to new tasks, thus featuring constraints in terms of modularity, to allow for reusability. Furthermore, policies need to be human readable, to allow for monitoring and debugging but also to make robot intentions understandable to close-by human operators.
\par
Behavior Trees (BTs) are a task switching policy representation with the properties of reactivity, readability and modularity and they are becoming state-of-the-art for robot control, especially in the research environment~\cite{iovino_survey_2022}. In industry however, Finite State Machines (FSMs) are still the preferred policy representation, due to their intuitive and simple design and a more mature exploitation in the field. 
\par
We believe that industry still needs to be provided with concrete proofs about the differences on the policy behavior in robotic tasks. The goal of this paper is to compare BTs with FSMs in terms of reactivity, modularity, readability, design choices, and more practically in a set of mobile manipulation tasks. For the comparison to be fair, we use the same low level implementation for the robot skills, changing only the encapsulating container to a behavior for a BT or a state for an FSM.

The main contribution of this paper is to provide a set of concrete examples that shows what these properties mean in robotic applications. We support the comparison with common metrics for algorithm complexity and graph distances that suggest the advantages of using BTs as an alternative to FSMs. The reader is referred to~\cite{GUGLIERMO2024104714} for a more in-depth analysis on metrics for evaluating BTs.
\par
This paper extends our previous work~\cite{iovino_programming_2022} with the following points:
\begin{enumerate}
    \item we compare BTs and FSMs on the reactivity and readability properties other than the modularity;
    \item we extend the comparison between BTs and FSMs on the modularity aspect by providing more examples;
    \item we extend the comparison by considering Hierarchical FSMs as well;
    \item we extend the experimental section by evaluating the metrics on concrete examples and by implementing the two policies on a real robotic system.
\end{enumerate}
\par
The remainder of this paper is organized as follows. Section~\ref{sec:work} provides background on the two policy representations and on related work that compares them. Section~\ref{sec:design} describes how policies can be automatically generated and presents the design choices that are considered for both BTs and FSMs. The metrics as well as the policy properties considered for the comparison are detailed in Section~\ref{sec:metrics}, while the practical realization is presented in Section~\ref{sec:experiments}. Finally, Section~\ref{sec:conclusion} summarizes the results of the comparisons.

\begin{figure}[tbp]
    \centering
    \includegraphics[width=.9\linewidth]{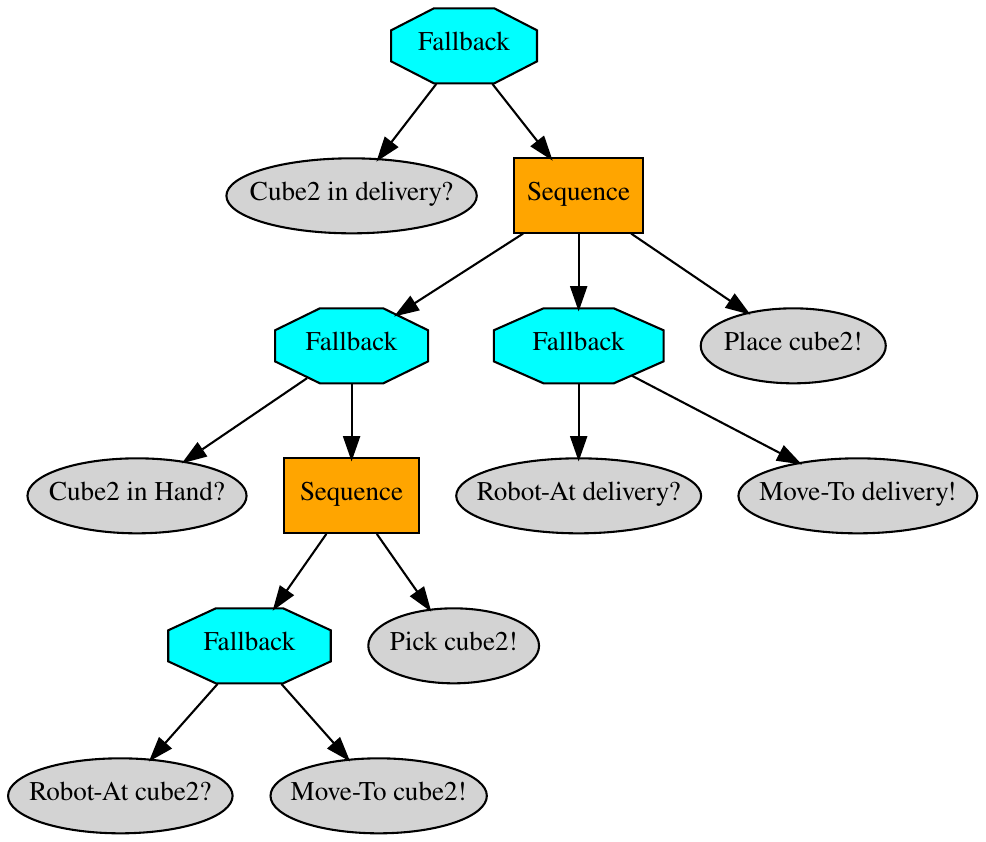}
    \caption{Behavior Tree automatically generated by a planner using the backchaining method.}
    \label{fig:base_BT}
\end{figure}

\section{Background and Related Work} \label{sec:work}

This section briefly defines BTs and FSMs and presents the theoretical foundations and early results on the comparison between the two representations.

\subsection{Behavior Trees}

Behavior Trees are a representation for task switching controllers that originated in the gaming industry but then rapidly gained success in robotics applications~\cite{colledanchise_behavior_2018, ghzouli2020behavior, rovida_extended_2017, paxton_costar_2017}.
\par
A BT is a directed tree that is recursively ticked with a depth-first pre-order traversal. Internal nodes are called \emph{control nodes}, (polygons in Fig.~\ref{fig:base_BT}), where the most common types are \emph{Sequence}: executing children in a sequence, returning once all succeed or one fails, \emph{Fallback} (or \emph{Selector}): executing children in a sequence but returning when one succeeds or all fail, and \emph{Parallel}: executing children in parallel, returning when a pre-determined subset of children is successful.
Leaves are called \emph{execution nodes} or \emph{behaviors} (ovals in Fig.~\ref{fig:base_BT}) and are either (i) Action nodes that execute a behavior when ticked and return one of the status signals \{\textit{Running}, \textit{Success}, \textit{Failure}\}, or (ii) Condition nodes that encode status checks and sensory feedback and immediately return \textit{Success} or \textit{Failure}. 
\par
BTs have explicit support for task hierarchy, action sequencing, and reactivity~\cite{iovino_survey_2022}. BTs are modular by design: since every node has the same return arguments, every subtree can be seen as a building block that can be moved around and reused without compromising the structure of the policy. Moreover, modularity ensures that every building block is independent and thus can be tested separately. Reactivity is ensured by the \emph{Running} return state, which allows the whole tree to be logically evaluated at every tick and running actions to be preempted if some behavior with higher priority needs to be executed.
BTs are functionally close to Decision Trees with the main difference in the \textit{Running} state that allows BTs to execute actions for longer than one tick.

\begin{figure}[tbp]
    \centering
    \includegraphics[width=.6\linewidth]{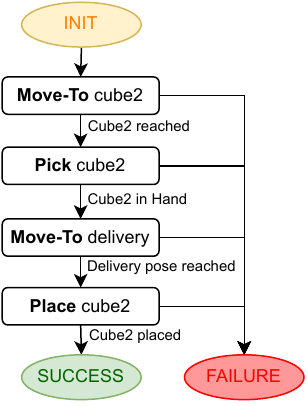}
    \caption{State Machine executing a sequence of actions to solve a mobile pick and place task.}
    \label{fig:seq_SM}
\end{figure}

\subsection{Finite State Machines}

Finite State Machines derive from state automata and feature a set of states and transitions between them (ovals and arrows in Fig.~\ref{fig:seq_SM}, respectively). Every state encodes a controller for robot behavior which produces effects in the environment upon execution. The effects trigger an event that transfers the execution from one state to the next. Since FSMs also include Sequential Function Charts (SFCs), a graphical programming language for Programmable Logic Controllers (PLCs), they are widely used in industry~\cite{lepuschitz_using_2019} and their success is mainly due to their intuitive design and implementation simplicity.
\par
FSMs have the unfortunate shortcoming that designers have to make a trade-off between reactivity and modularity.
As highlighted in~\cite{colledanchise_behavior_2018}, FSM execution can be compared to the GoTo statement of early programming languages, where the execution flow jumps from one part of the program to another and continues from there. In programming, and by extension also in robotics, GoTo statements are considered to be harmful, as reported in~\cite{dijkstra_letters_1968} by Dijkstra. On the other end, the execution of a BT can be compared to a function call, where the execution flow also jumps to another part of the code but after its completion, it returns to where the function was initially called.
\par
In order to be reactive, an FSM needs to have many transitions that need to be taken care of upon addition or removal of states, making them less modular and less scalable. The modularity problem is partially mitigated by logically grouping states to form hierarchies. Even if more reusable, it is still hard to add or remove states in Hierarchical Finite State Machines (HFSMs) and often hierarchies are handcrafted, actually shifting the modularity problem to inner layers. Reactive HFSMs often result in fully connected graphs~\cite{colledanchise_behavior_2018}. 

Modularity in FSMs can also be achieved by parallel composition~\cite{baier2008principles}. With this operator it is possible to design FSMs for different subtasks independently and then use the parallel composition to automatically generate an FSM for the complete task. With parallel composition, in the generated FSM the number of states grows combinatorially with the states in the sub-FSMs that compose it. Therefore, this operator grants modularity at the cost of higher complexity and loss of readability. Due to this complexity, parallel composed FSMs are often represented symbolically and not graphically.

\subsection{Related Work}

BTs have been compared to FSMs in previous work, but the comparison was either purely theoretical~\cite{colledanchise_behavior_2018, colledanchise_how_2017, biggar_expressiveness_2021}, or speculative~\cite{klockner_behavior_2013}. In~\cite{colledanchise_behavior_2018}, the authors list advantages and disadvantages of both designs and in~\cite{colledanchise_how_2017} they prove theoretically how in fact BTs modularize FSMs, providing a design of an HFSM that behaves like a BT. In~\cite{biggar_expressiveness_2021}, the authors compare BTs with other related architectures (Decision Trees, Teleo-reactive Programs and in particular FSMs) in terms of reactivity, readability and expressiveness. They demonstrate that \textbf{FSMs are more expressive than BTs} since they can build behaviors based on actions that have access to internal variables and to past decisions, while \textbf{BTs are more reactive} because they restart the execution from the root at each new input \textbf{and more readable} as they do not encode past decisions in the representation.

In~\cite{biggar_modularity_2021}, the same authors formalize modularity for reactive control architectures, pointing out that BTs feature structural interfaces, with which \textit{every component} interacts with the others: a subtree is a BT, an action behavior is a degenerate case of a BT. Naturally, FSMs lack structural interfaces and thus cannot be considered modular in their analysis unless a structure is enforced (e.g. in HFSMs). Modularity is measured by the Cyclomatic Complexity, defined as 
\begin{equation}
    CC = a+s-n+1
\end{equation}
where $a, s, n$ represent the number of arcs, sinks (terminal nodes) and nodes in a decision structure, respectively. This measure is applied to BTs that have been transformed to graphs with single entry and exit nodes. This implies that \textbf{BTs have optimal modularity} as their Cyclomatic Complexity is 1.
In addition to this analysis, we measure modularity in terms of the efforts required to modify a structure by adding or removing elements in it. We propose to quantify using the computational complexity of such operations and the edit distance between a baseline structure and its modified versions. 
\par
Kl\"ockner~\cite{klockner_behavior_2013} proposes using a BT as a control policy for UAV missions, speculating the advantages of such design with respect to FSMs.
Previous work~\cite{olsson_behavior_2016} also compares the two policy representations from a practical perspective in the domain of autonomous driving. The comparison is made in terms of Cyclomatic Complexity (CC) and Maintainability Index (MI). The MI of a piece of software takes into account the CC, the number of lines of code, the percentage of comments in the code and the Halstead volume (a function of distinct and total numbers of operands and operators). Here, the CC was ill-defined as compared to~\cite{biggar_modularity_2021} and we argue that the MI is more dependent on the library used to implement the two policies, so we disregard it.
We propose instead to count the number of elements in the structure. We are aware that this approach is not a metric per se, but we use it to indicate how the structural complexity evolves with respect to the task complexity. 
\par
Another work that provides insight on the behavior of an agent controlled by both policies is~\cite{colledanchise_learning_2018}, in the domain of computer games, i.e. in a deterministic scenario. Here they evaluate the policies on solving a level in the Mario AI benchmark, comparing them in terms of number of nodes and reward function $\rho(x)$. This comparison provides insight on the scalability of the two policies, since both of them are generated by a learning algorithm to solve the benchmark. The goal of this work was to propose a new method to generate BTs, so the comparison against FSM is not fair in the sense that the generation method is not the same for the two policies.
Nevertheless, from both~\cite{olsson_behavior_2016} and~\cite{colledanchise_learning_2018} we can conclude that \textbf{the complexity scales linearly for BTs and quadratically or worse for FSMs}.
\par
Authors in~\cite{wuthier_productive_2021} instead, propose a combined design where some nodes of the BT are modelled as FSMs. The authors claim that with such design the FSMs allow the policy to easily define states with memory which would instead be modelled in BTs by conditions with complex semantics. A similar combined design, where instead states in an FSM are represented as BT, is proposed in~\cite{hallen2024behavior}. Having an FSM at the highest level of the representation, allowed human users to control the switching between modes, while the modularity of BTs makes it preferable for the actual robot behavior implementation. A similar hybrid model is implemented in~\cite{zutell_flexible_2022} where BTs are combined with HFSMs.

\par
Finally, authors in~\cite{caceres_dominguez_stack--tasks_2022} combine BTs with Stack-of-Tasks (SoT) control, a paradigm that allows the robot to fulfill a list of goals formulated as equality or inequality constraints in error space. Such goals have a priority order and can be fulfilled simultaneously. SoT is usually combined with FSM to prevent the robot from getting stuck in local minima. However, authors in~\cite{caceres_dominguez_stack--tasks_2022} claim that combining SoT with BTs instead solves limitations of the FSMs in terms of reactivity, modularity and re-usability.

\section{Policy Design} \label{sec:design}

To allow for a fair comparison as well as a fast implementation, we selected policy implementations that are based on \texttt{python} and are ROS-compatible. For BT design we use \texttt{py\_trees}\footnote{\url{https://github.com/splintered-reality/py_trees}}, while for FSM we use SMACH~\cite{bohren_smach_2010}\footnote{\url{https://github.com/ros/executive_smach}}.
\par
There exist several methods for automatically generating Behavior Trees, such as Genetic Programming~\cite{iovino_learning_2021}, Learning from Demonstration~\cite{french2019learning, gustavsson_combining_2022}, autonomous planners~\cite{colledanchise_towards_2019}, or hybrid combinations~\cite{mayr2022combining, styrud_combining_2022, iovino_framework_2023}. Some of these generation methods build a BT using a backward-chained approach~\cite{colledanchise_towards_2019, gustavsson_combining_2022, iovino_framework_2023}, which is proven to have convergence guarantees~\cite{ogren_convergence_2020}. For this method, actions are defined together with their pre- and post-conditions. Starting from the goal, pre-conditions are iteratively expanded with actions that achieve them, i.e. those actions that have that particular condition as one of their post-conditions. Then, those actions' unmet pre-conditions are expanded in the same way. Another property of backchained BTs, is that the control node types are in an alternating order, i.e. the parent of a \textit{Fallback} node is a \textit{Sequence} node and vice-versa, resulting in a design recommended as good practice in~\cite{colledanchise_behavior_2018}. By providing a planner with such a set of actions and a goal state, it is possible to automatically generate backchained BTs (Fig.~\ref{fig:base_BT}).

\begin{figure}[tbp]
    \centering
    \includegraphics[width=.9\linewidth]{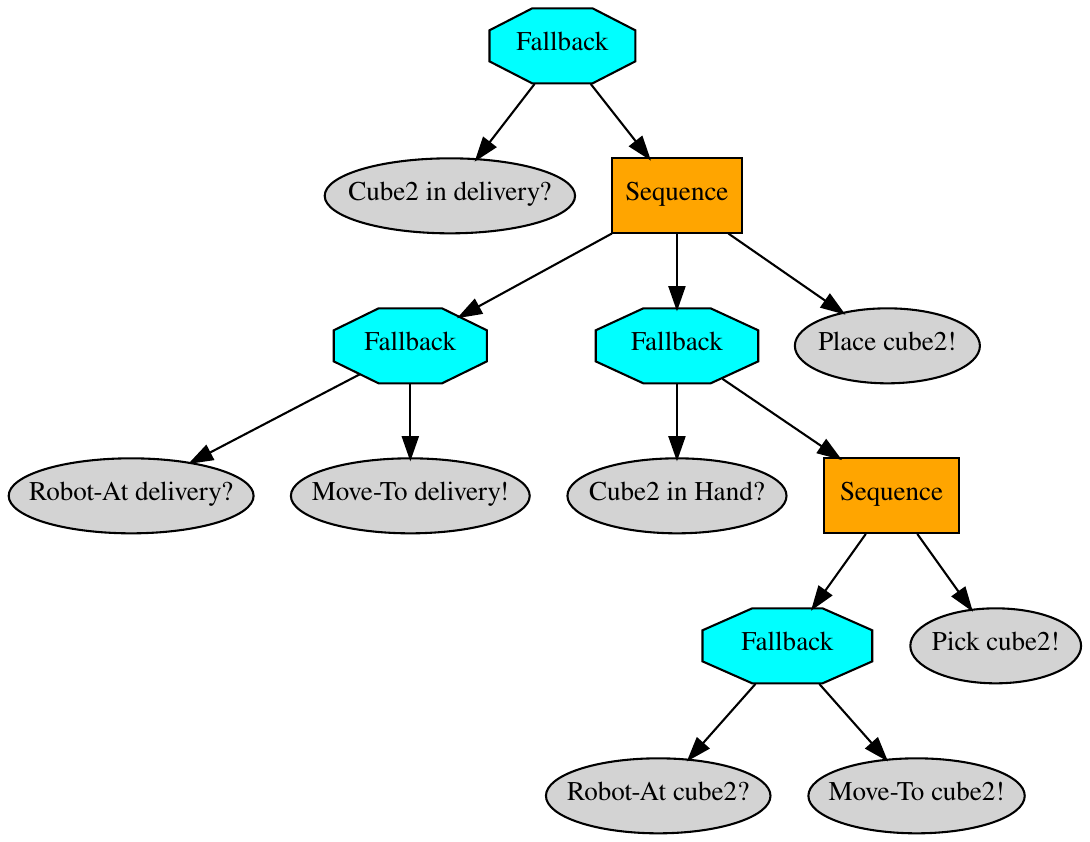}
    \caption{Backchained BT realizing a chattering controller.}
    \label{fig:chatter_BT}
\end{figure}

Note that in BT design, the order of the actions' pre-conditions is important. In the case of the BT in Fig.~\ref{fig:base_BT}, for example, if the pre-condition \textit{`Robot-At delivery?'} of the \textit{`Place cube2!'} action is expanded before the pre-condition \textit{`Cube2 in Hand?'}, we would obtain the BT of Fig.~\ref{fig:chatter_BT}. A robot controlled with such a BT will feature a chattering behavior because the robot will first navigate to the place position (the condition \textit{`Robot-At delivery?'} will then return \emph{Success}), then it will attempt picking the cube, by navigating to the picking pose with the action \textit{`Move-To cube2!'}. As soon as the robot moves, the condition \textit{`Robot-At delivery?'} will immediately return \emph{Failure}, and by consequence re-execute the action \textit{`Move-To delivery!'}. This chattering behavior will continue indefinitely unless there is a limit on the number of ticks allowed. To avoid this problem, the order of the pre-conditions is compared against the order in which actions are executed in the plan, so the pre-condition \textit{`Cube2 in Hand?'} for the picking action is expanded before the pre-condition \textit{`Robot-At delivery?'} for the moving action.
\par
Another degree of freedom in backchained BT design is the order of actions for a given post-condition. In this case, a single post-condition would be achieved by different actions, resulting in different subtrees when those actions are expanded. A good design policy would be to sort the subtrees according to e.g. execution time, success probability or some a priori defined cost. In this direction, authors in~\cite{fusaro_human-aware_2021} assign a cost to every action and at runtime the robot executes the subtree with the lowest cost.
\par

\begin{figure}[tbp]
    \centering
    \includegraphics[width=.9\linewidth]{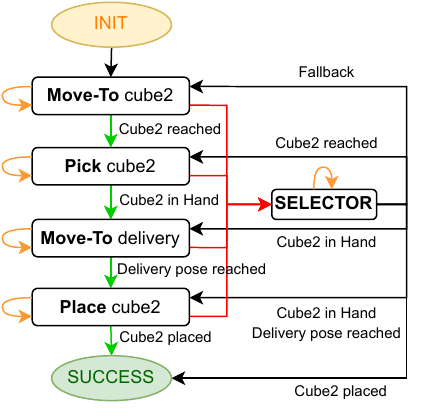}
    \caption{Reactive State Machine to solve a mobile pick and place task. Transitions are \textit{Running} (yellow), \textit{Success} (green) and \textit{Failure} (red).}
    \label{fig:fault_SM}
\end{figure}

\begin{figure}[tbp]
    \centering
    \includegraphics[width=.6\linewidth]{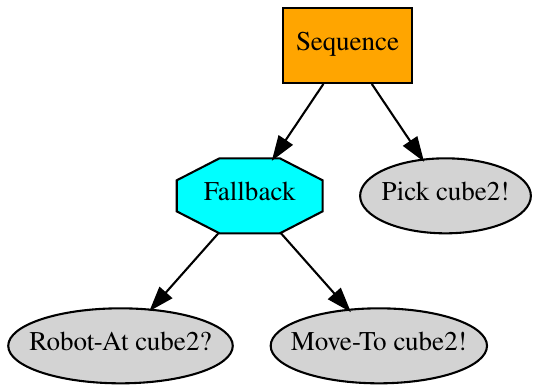}
    \caption{Example BT subtree. Its equivalent HFSM is shown in Fig.~\ref{fig:HFSM}.}
    \label{fig:subtree}
\end{figure}

\begin{figure}[tbp]
    \centering
    \includegraphics[width=.75\linewidth]{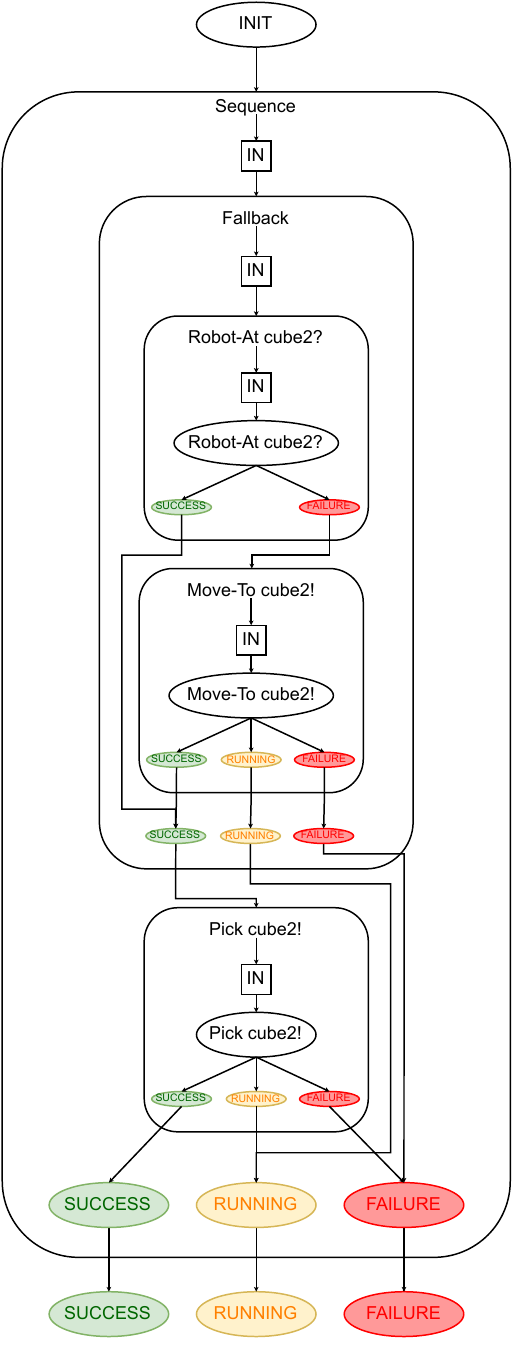}
    \caption{HFSM that mimics the subtree in Fig.~\ref{fig:subtree}.}
    \label{fig:HFSM}
\end{figure}

\begin{table*}[!ht]
\centering
\caption{Overview on the performances of the policies according to the metrics. In a BT $n$ is the number of nodes, $n^{\star}$ is the number of nodes to modify in an edit operation, and $M$ the number of action nodes. In an FSM $n$ is the number of states and transitions, $n^{\star}$ is the number of states to modify in an edit operation, while $M$ is the number of states alone. In particular we use the following notation $T_{fc} = M_{fc}(M - 1)$ where $M_{fc}$ is the number of fully connected states in an FSM.}
\begin{tabular}{c c c c c c}
& \multicolumn{5}{c}{\fontsize{12}{0}\selectfont \textbf{Metrics}}\\
\cmidrule{2-6}
& \multicolumn{2}{c}{\fontsize{10}{0}\selectfont \textbf{Modularity}} & \multicolumn{1}{c}{\fontsize{10}{0}\selectfont \textbf{Reactivity}} & 
\multicolumn{2}{c}{\fontsize{10}{0}\selectfont \textbf{Readability}} \\
\cmidrule{2-6}
\multicolumn{1}{c}{\fontsize{12}{0}\selectfont \textbf{Policy}} & 
\multicolumn{1}{c}{Computational Complexity} & 
\multicolumn{1}{c}{ Edit Distance} & 
\multicolumn{1}{c}{Effort} & 
\multicolumn{1}{c}{Graphical Elements} &
\multicolumn{1}{c}{Active Elements} \\
\cmidrule(r){1-1} \cmidrule{2-6}
\multicolumn{1}{r}{Backchained BTs}
 & $\mathcal{O}(1)$ & $2n^{\star}$ & $0$ &  $\simeq 7M-1$ & $\simeq 3.5M$  \\
\cmidrule(lr){2-6}
\multicolumn{1}{r}{Fault-Tolerant FSMs}
 & $\mathcal{O}(n)$ & $f(n^{\star}, n)$ & $3(M + 1) +T_{fc}$ &  $\simeq 5M+4 + T_{fc}$ & $\simeq 5M+4 + T_{fc}$ \\
\cmidrule(lr){2-6}
\multicolumn{1}{r}{HFSMs}
 & $\mathcal{O}(1)$ & $kn^{\star}$ & $0$ & $\simeq 36M-3$ & $\simeq 29M-3$ \\
\bottomrule
\end{tabular}
\label{tab:properties}
\end{table*}

With those inputs the design of a BT is somewhat constrained, while there are multiple design choices for an FSM. A first option is to directly mimic the sequence of actions, as depicted in Fig.~\ref{fig:seq_SM}. This design has the shortcoming that it is not reactive: a robot controlled by such a policy will not react to events that disrupt the task at any given time, nor is it possible for the robot to recover the execution at any point of the task but only by returning to the beginning. To overcome this limitation, it is necessary to turn the FSM into a fully connected graph, taking into account all failure cases, or more simply, to connect every state to a SELECTOR (or failure) state where the state of the robot and the environment is inspected before the next state is reached (Fig.~\ref{fig:fault_SM}). In this case, the FSM features a \emph{Running} transition that cycles back to the same state, allowing the FSM to preempt an action execution provided the execution is asynchronous and the environment is monitored periodically. Then, every execution state has a \emph{Failure} transition to the SELECTOR state, so that the execution can restart in the appropriate state, depending on the task progression. In case the FSM is generated by a planner, the transition from the SELECTOR state to another state is triggered by that state's pre-condition. The other transitions are the same as in the previous case and depending on the execution sequence.
\par
Another design alternative is the one proposed in~\cite{colledanchise_how_2017}. In this work, the authors formulate an FSM as a Hierarchical Finite State Machine (HFSM), with the characteristic that it behaves exactly as a BT. Every robot behavior is an FSM that features the states \emph{Running, Success, Failure} as outcomes; two or more behaviors are encapsulated in another FSM and transitions are made according to whether the parent FSM mimics a \emph{Sequence} or a \emph{Fallback} node. An example is provided in Fig.~\ref{fig:HFSM}, where the HFSM mimics the subtree in Fig.~\ref{fig:subtree}. In the \emph{Fallback} container, if the condition is successful, the execution is transferred directly to the \textit{Success} outcome of the container, or to the next action in case of failure. In the higher level \emph{Sequence} container, if the \emph{Fallback} container is successful, then the execution is transferred to the next action, or to a \textit{Failure} outcome otherwise.

\section{Metrics-based Comparison} \label{sec:metrics}

In this section we compare Behavior Trees and Finite State Machines in terms of modularity, reactivity, readability and we conclude it with comments on alternative designs. We will describe each property separately and provide dedicated metrics where applicable, the results are summarized in Table~\ref{tab:properties}.

\begin{figure*}[tbp]
     \centering
\begin{subfigure}[b]{0.5\textwidth}
    \centering
    \includegraphics[width=\linewidth]{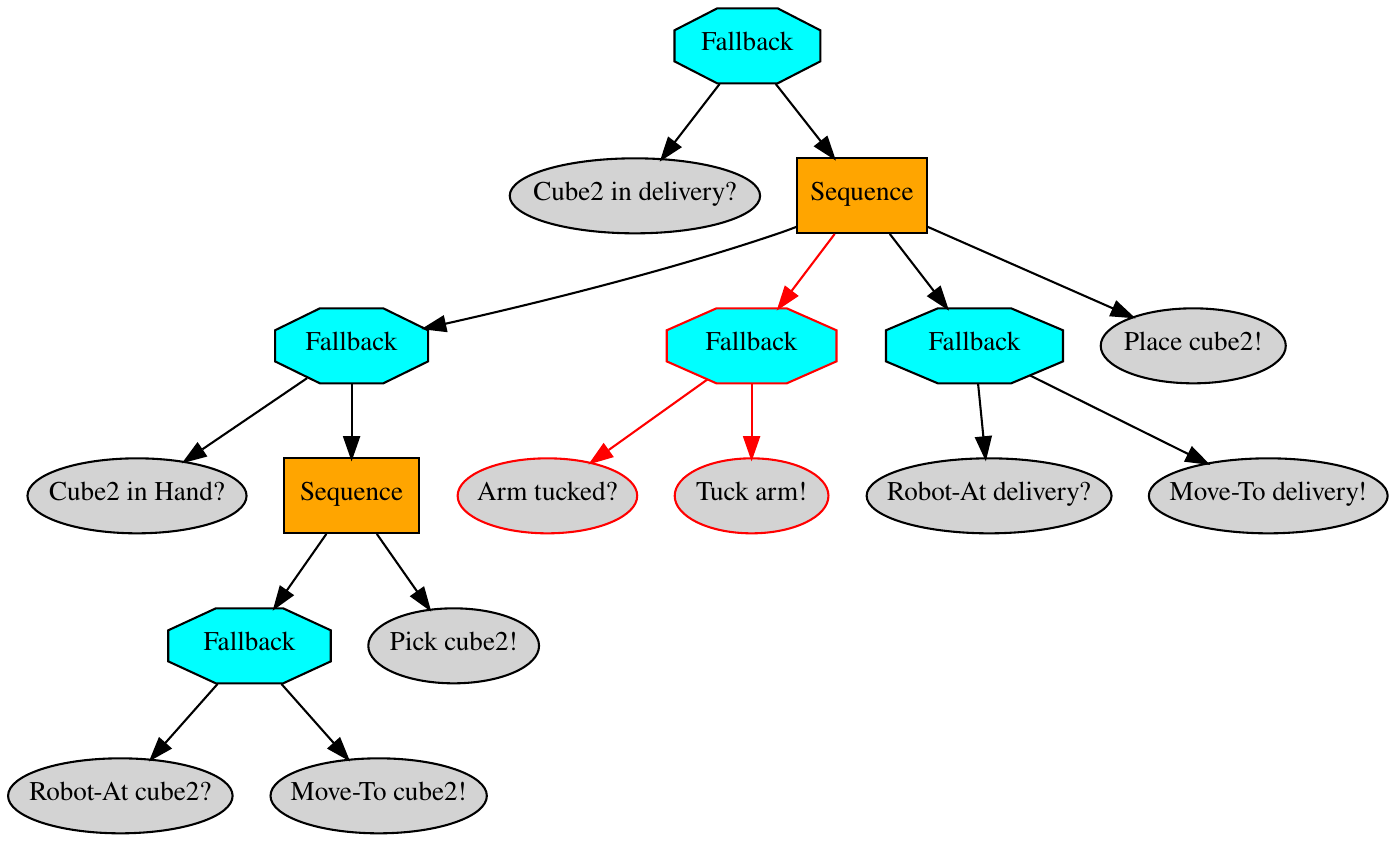}
    \caption{Adding the subtree to tuck the arm in the BT: the robot will tuck the arm after grasping.}
    \label{fig:tuck_BT}
\end{subfigure}
\hfill
\begin{subfigure}[b]{0.45\textwidth}
    \centering
    \includegraphics[width=\linewidth]{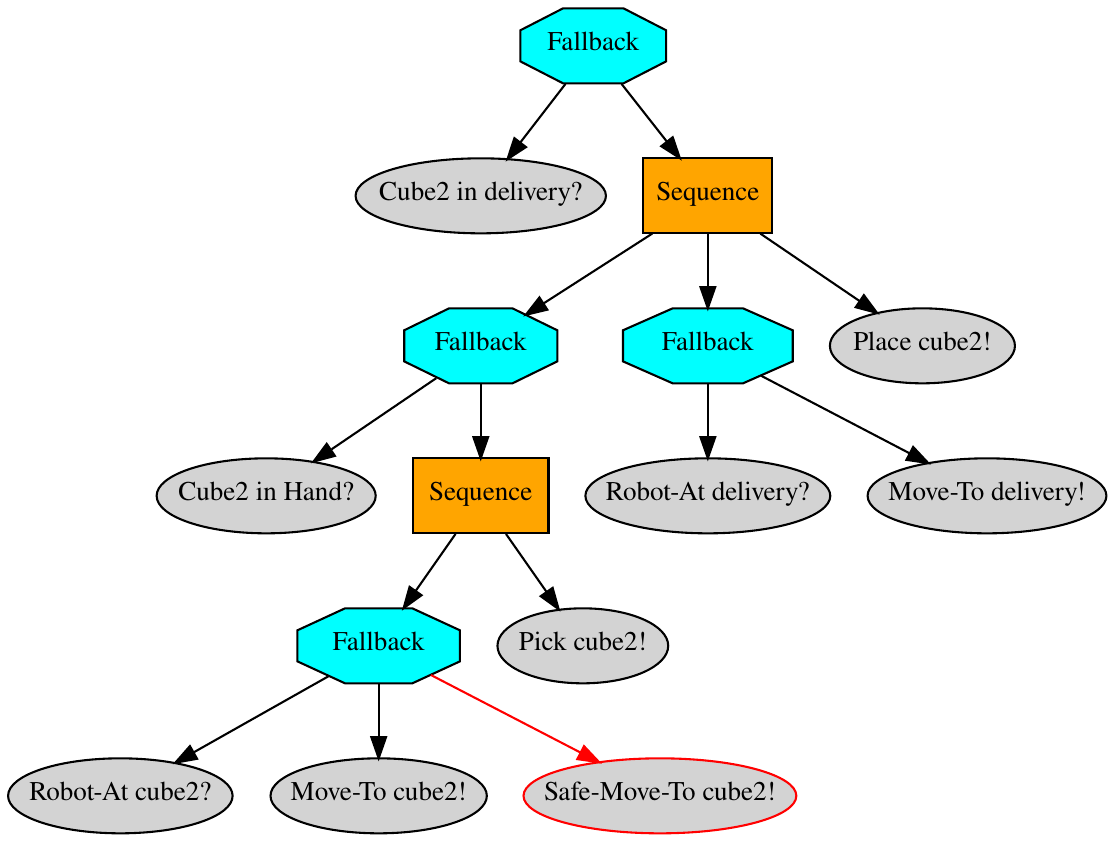}
    \caption{Adding an alternative way to move to the target cube: if the first Move-To behavior fails, a safer motion is attempted.}
    \label{fig:safer_BT}
\end{subfigure}
\hfill
\begin{subfigure}[b]{0.45\textwidth}
    \centering
    \includegraphics[width=\linewidth]{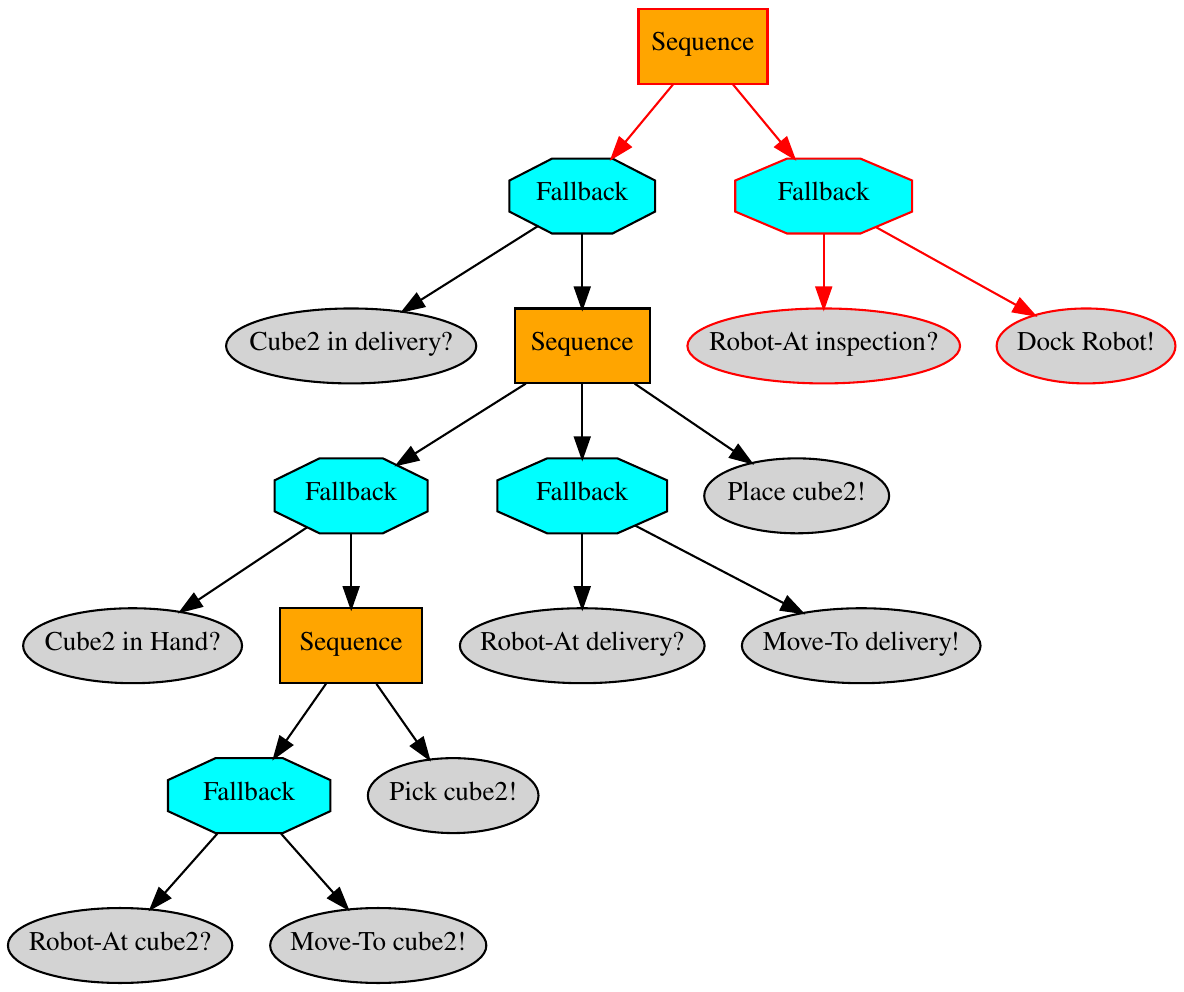}
    \caption{Adding a final behavior for the task in a BT.}
    \label{fig:dock_BT}
\end{subfigure}
\hfill
\begin{subfigure}[b]{0.5\textwidth}
    \centering
    \includegraphics[width=\linewidth]{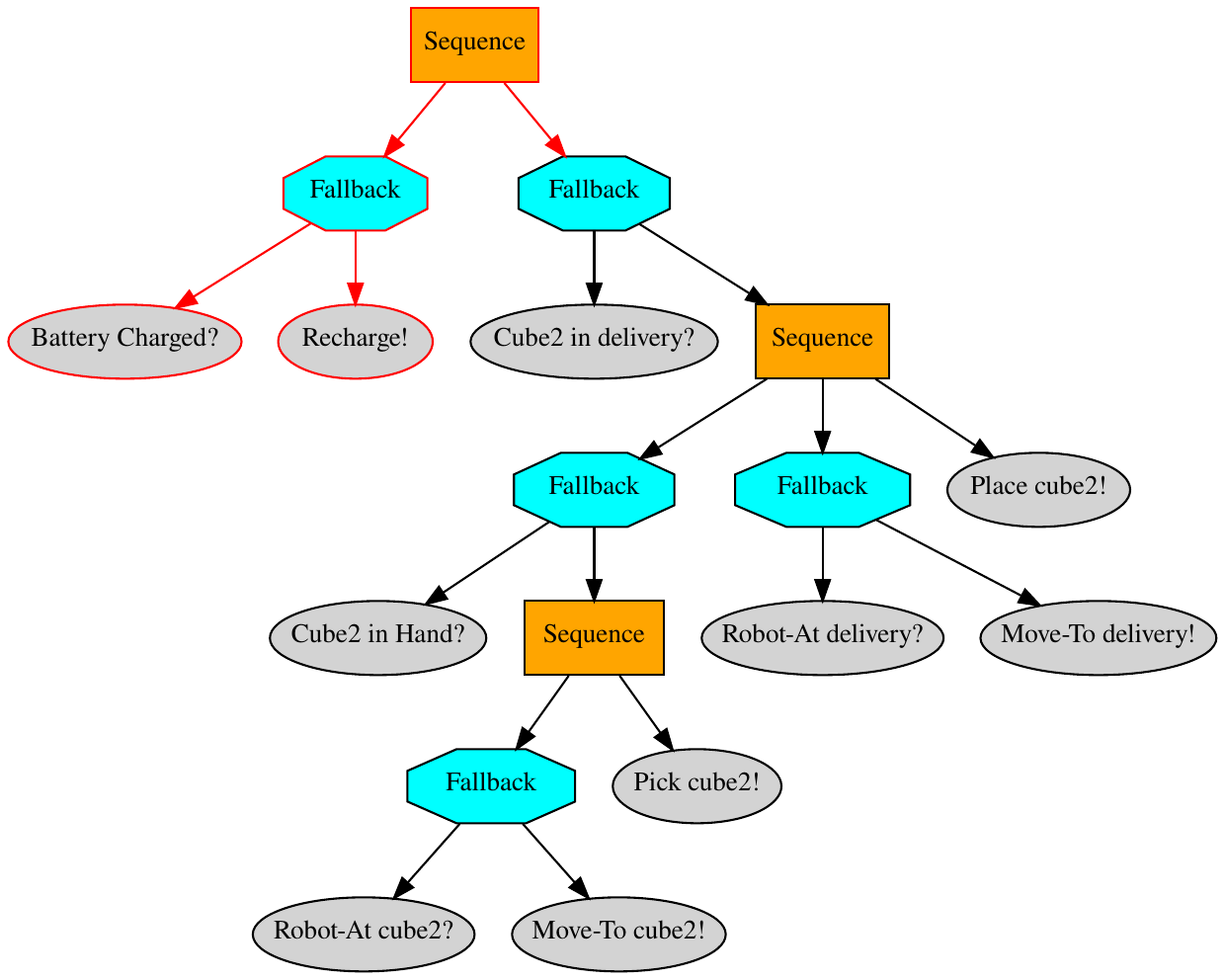}
    \caption{A BT that charges the robot if it is low on battery.}
    \label{fig:battery_BT}
\end{subfigure}
\hfill
\begin{subfigure}[b]{0.7\textwidth}
    \centering
    \includegraphics[width=\linewidth]{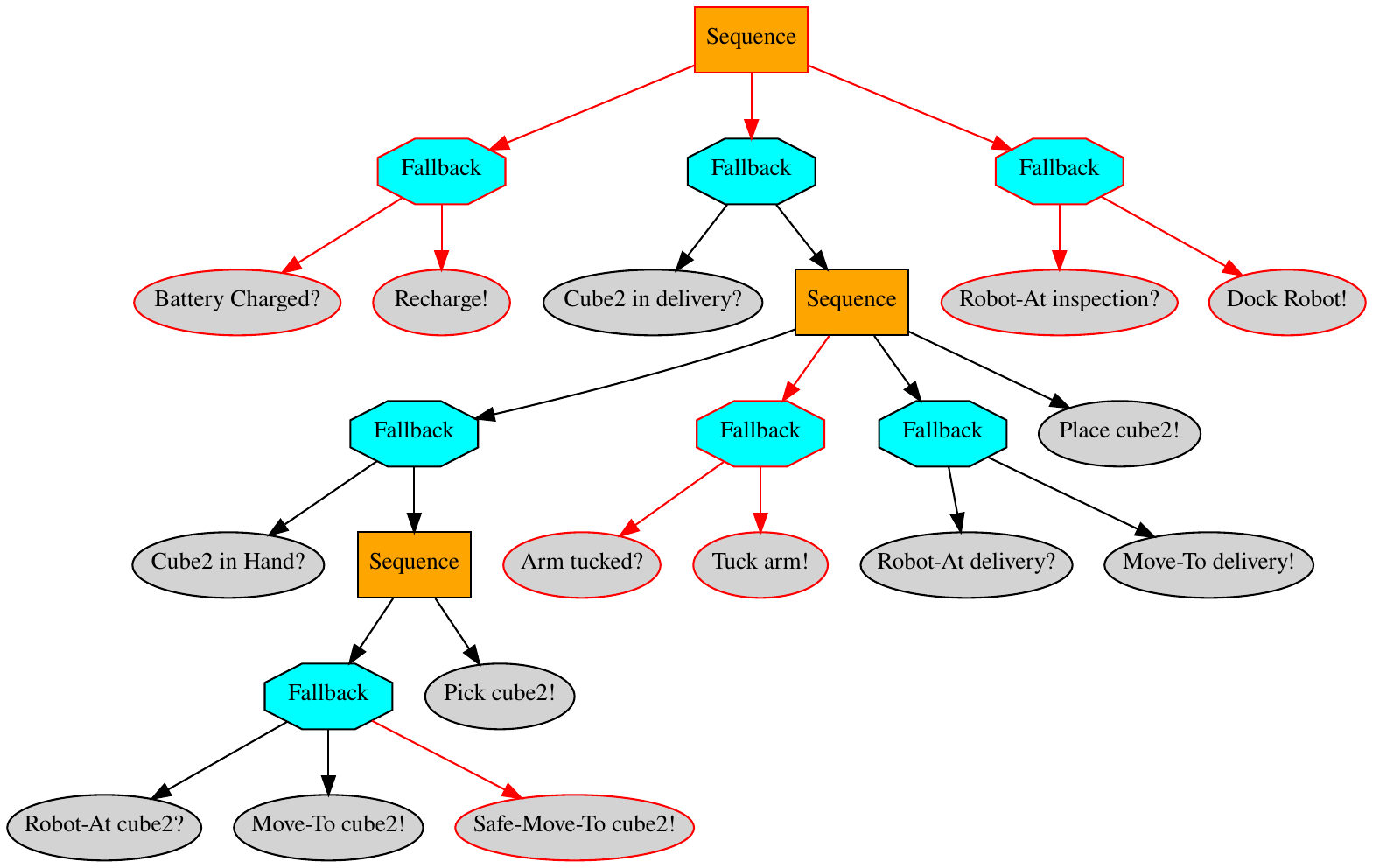}
    \caption{BT resulting from all previous node addition cases.}
    \label{fig:addall_BT}
\end{subfigure}
\caption{Different cases of node addition in a BT. The added nodes are highlighted in red.}
        \label{fig:add_BT}
\end{figure*}

\begin{figure*}[tbp]
     \centering
\begin{subfigure}[b]{0.38\textwidth}
    \centering
    \includegraphics[width=\linewidth]{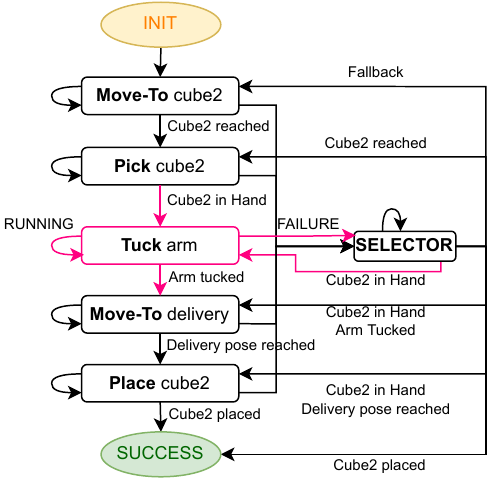}
    \caption{Adding an intermediate behavior for the task in an FSM: the robot tucks the arm after grasping.}
    \label{fig:tuck_SM}
\end{subfigure}
\hfill
\begin{subfigure}[b]{0.51\textwidth}
    \centering
    \includegraphics[width=\linewidth]{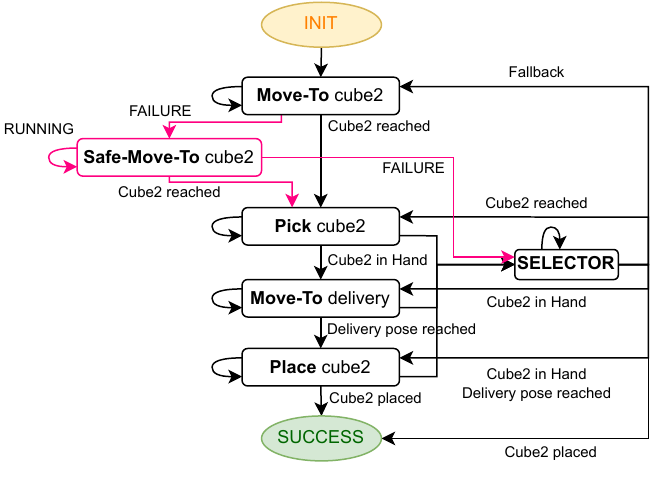}
    \caption{Adding an alternative way to move to the target cube: if the first Move-To behavior fails, a safer motion is attempted.}
    \label{fig:safer_SM}
\end{subfigure}
\hfill
\begin{subfigure}[b]{0.38\textwidth}
    \centering
    \includegraphics[width=\linewidth]{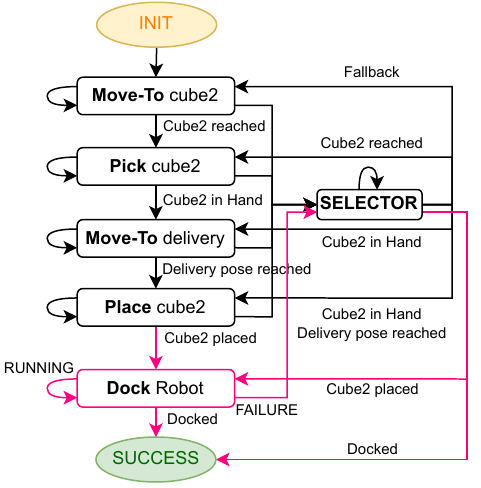}
    \caption{Adding a final behavior for the task in an FSM: the robot docks before finishing the task.}
    \label{fig:dock_SM}
\end{subfigure}
\hfill
\begin{subfigure}[b]{0.51\textwidth}
    \centering
    \includegraphics[width=\linewidth]{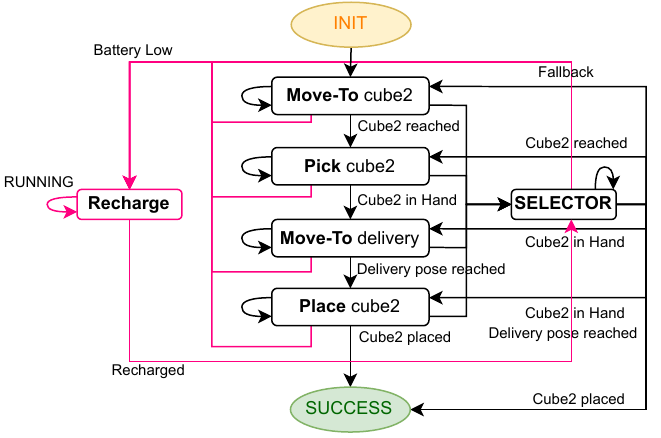}
    \caption{Adding a connected behavior for the task in an FSM: the robot charges the batteries if low.}
    \label{fig:battery_SM}
\end{subfigure}
\hfill
\begin{subfigure}[b]{0.47\textwidth}
    \centering
    \includegraphics[width=\linewidth]{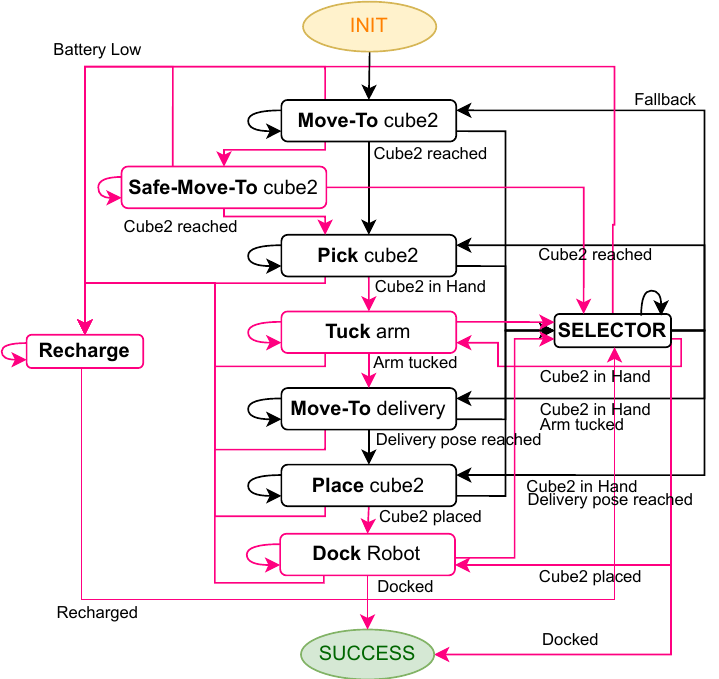}
    \caption{FSM resulting from all previous node addition cases.}
    \label{fig:addall_SM}
\end{subfigure}
\caption{Different cases of state addition in a fault-tolerant FSM. The added elements are displayed in magenta.}
        \label{fig:add_SM}
\end{figure*}

\subsection{Modularity} \label{sec:modularity}

The possibility of adding, removing, and editing nodes/states, allows the users to reuse pieces of software and to modify bad design choices. It can happen, for example, that during the testing of a policy the user realizes that some behaviors were missing or that some were malfunctioning, and thus need to be removed or replaced. In this section, we will compare the addition, removal, and modification of behaviors in the two decision structures with a focus on the computational complexity of such operations and on the edit distance (a measure of diversity between two graphs) between a baseline structure and the modified one. To do so, we will use the baseline structures of Figs.~\ref{fig:base_BT} and~\ref{fig:fault_SM} and we modify them according to the following scenarios:
\begin{enumerate}
    \item we add to the baseline a behavior that tucks the robot arm after grasping, to have an arm configuration more suited for navigation;
    \item we add an alternative way of reaching the first item to pick, by letting the robot generating a safer trajectory that allows it to move closer to obstacles but with a reduced velocity. This behavior is used when the standard Move-To behavior fails;
    \item once the mobile manipulation task is finished, we want the robot to reach a docking station;
    \item we make the robot recharge the batteries if they run low at any point during task execution.
\end{enumerate}

The structures that we obtain by modifying the baseline are grouped in Fig.~\ref{fig:add_BT} in the case of BTs and Fig.~\ref{fig:add_SM} for FSMs.

\subsubsection{\textbf{Computational Complexity}}

\paragraph{Adding or removing a node in a BT} Adding a node in a BT with $n$ nodes, requires inserting it in the list of children of the parent control node at the desired position. In a BT, every child node is in no way connected to other children\footnote{This is not true in case we use Blackboard variables, but this design choice breaks modularity and it is discouraged~\cite{colledanchise_behavior_2018, shoulson_parameterizing_2011}.}.  Performing the insertion operation requires accessing the desired parent node and then inserting it, with complexity $\mathcal{O}(1)$. Removing a node has the same complexity.
Because of modularity, handling a subtree has the same complexity. In this case, the root of the subtree to add/remove is considered as the input node.

There are special cases in which we might desire to prepend or append a subtree to modify the behavior of the robot. For example, if we want the robot to recharge the batteries as soon as they run low, we would need to prepend this subtree in a \emph{Sequence} node root, to the subtree solving the task (so the recharging subtree has higher priority). The BT obtained this way is depicted in Fig.~\ref{fig:battery_BT}, where the added elements are highlighted in red. In this particular case, a \textit{Sequence} root node is created, then the subtrees for recharging the robot and solving the mobile manipulation tasks are added as children to the root. Similarly, we can append another subtree to the root, as in Fig.~\ref{fig:dock_BT}, docking the robot after the manipulation task.

\paragraph{Adding or removing a state in a FSM} For the proposed design of the FSM, we identify three possible ways a state can be added:
\begin{enumerate}
    \item[1)] \textbf{sequential state}: add a new state between two existing states (namely the \textit{preceding} and the \textit{following} state). If we want to add the new state as a new step in the execution sequence, we need to remove the transition from the \textit{preceding} state to the \textit{following} one, then to create a transition from \textit{preceding} to the new state and from this to the \textit{following}. Finally we need to handle the interaction with the SELECTOR state. An FSM modified this way is reported in Fig.~\ref{fig:tuck_SM}. If the new state is a terminal one, then we would make a transition to the outcome instead. An FSM modified this way is reported in Fig.~\ref{fig:dock_SM}.
\end{enumerate}
\begin{enumerate}
    \item[2)] \textbf{alternative state}: add a new state as an alternative strategy for an existing one. In this case, the \textit{FAILURE} transition of the \textit{preceding} state goes to the new one instead of SELECTOR. A copy of the transition from \textit{preceding} to \textit{following} is used also from the new state to the \textit{following}. Finally, a \textit{FAILURE} transition is added from the new state to SELECTOR. An FSM modified this way is reported in Fig.~\ref{fig:safer_SM}.
\end{enumerate}
\begin{enumerate}
    \item[3)] \textbf{connected state}: add a new state for the task that is connected to all the others (Algorithm~\ref{alg:sm_add3}). In this case, we need to add a transition from every other state to the new one and finally to handle the interaction with the SELECTOR state. An FSM modified this way is reported in Fig.~\ref{fig:battery_SM}.
\end{enumerate}

\begin{algorithm2e}[!ht]
\caption{Adding a connected state in an FSM.}\label{alg:sm_add3}
\SetStartEndCondition{ }{}{}%
\SetKwProg{Fn}{def}{\string:}{}
\SetKw{KwTo}{in}\SetKwFor{For}{for}{\string:}{}%
\SetKwIF{If}{ElseIf}{Else}{if}{:}{elif}{else:}{}%
\SetKwFor{While}{while}{:}{fintq}%
\AlgoDontDisplayBlockMarkers\SetAlgoNoEnd\SetAlgoNoLine%
\SetKwData{Cond}{condition}
\SetKwData{ICond}{selector\_condition}
\SetKwData{SELECTOR}{SELECTOR}
\SetKwData{State}{state}
\SetKwData{FSM}{sm}
\SetKwFunction{Outcome}{register\_outcome}
\SetKwFunction{AddTr}{add\_trans}
\SetKwFunction{AddSt}{add\_state}
\SetKwInOut{Input}{input}
\Input{sm, new\_state, condition, selector\_condition}
\BlankLine
\For{state \KwTo \FSM}{
    $state$.\Outcome(new\_state)

    \If{state == \SELECTOR}{
        $state$.\AddTr(\ICond)
    }
    \Else{
        $state$.\AddTr(\Cond)
    }
}
$new\_trans \gets ``RUNNING":\ new\_state$

$new\_trans \gets ``FAILURE":\ $\SELECTOR

\FSM.\AddSt($new\_state$, $new\_trans$)
\end{algorithm2e}

In an FSM with $n$ states, adding or removing a new state requires checking the consistences between state outcomes and transitions, which has complexity $\mathcal{O}(n)$ because we might need to take care of up to $n$ transitions.
\par
In particular, to remove a state, it is necessary to delete all transitions to and from the state. Then, the target state has to be removed from the outcomes of any other state, if present. This operation requires looping through all transitions and all states, for a total complexity of $\mathcal{O}(2n)$. Moreover, since the switching logic in an FSM is implemented inside the state and not explicitly realized by the representation, edit operations require the modification of the program script of the states. For example, if we want to remove a fully connected state, we need to remove or change the transition implementation in all states that reference it. 
\par
A fundamental difference with BTs\----a direct consequence of BT modularity\----is that while editing an FSM it is necessary to have access to all states and transitions, making the operation of adding/removing an element in the structure dependent on all the others. Moreover, we have seen that the three insertion scenarios have to be treated slightly differently. These problems are mitigated by using an HFSM.
In this case, adding a state has the same complexity as in the BT case. Once the insertion point is identified, the same \texttt{add\_state} method as for the FSM is called (Algorithm~\ref{alg:sm_add3}). However, here the consistency is checked locally (i.e., at a specific depth in the hierarchy) so the operation has a complexity $\mathcal{O}(k)$, with $k$ the number of children of the parent node at the insertion point. For example, if we were to insert a subtree to tuck the arm after grasping, as for the BT in Fig.~\ref{fig:tuck_BT}, the \texttt{add\_state} method would check for consistency only among the direct children of the parent \emph{Sequence} (i.e., the two \emph{Fallback} nodes and the place action). Transitions would also be handled locally and in a modular way, because we enforced a structure (i.e., three return statuses for every node). However, we trade off modularity for readability, as it can be appreciated by comparing Fig.~\ref{fig:subtree} and~\ref{fig:HFSM}.
\par


\begin{table}[!ht]
\centering
    \caption{Graph Edit Distance of the modified BTs in Fig.~\ref{fig:add_BT} to the baseline of Fig.~\ref{fig:base_BT} and the FSMs of Fig.~\ref{fig:add_SM} to the baseline of Fig.~\ref{fig:fault_SM}.}
\begin{tabular}{l c c c }
\multicolumn{4}{c}{\fontsize{12}{0}\selectfont \textbf{Graph Edit Distance}}\\
\midrule 
\hspace{.6cm} \fontsize{10}{0}\selectfont  \textbf{Modification} & \fontsize{10}{0}\selectfont \textbf{BT} & \fontsize{10}{0}\selectfont \textbf{FSM} & \fontsize{10}{0}\selectfont \textbf{HFSM} \\
\midrule
Tuck Arm subtree & 6 & 5 & 12 \\
\cmidrule(lr){1-4}
Safe-Move-To behavior & 2 & 4 & 4 \\
\cmidrule(lr){1-4}
Dock subtree & 8 & 5 & 17 \\
\cmidrule(lr){1-4}
Recharge Battery subtree & 8 & 8 & 17 \\
\bottomrule
\end{tabular}
\label{tab:graph_edit_dist}
\end{table}

\vspace{.5cm}
\subsubsection{\textbf{Edit Distance}}

The edit distance is a way to measure diversity between structures. With the formulation proposed in~\cite{burke_diversity_2004}, two BTs to compare are padded with empty nodes into the same shape. 


Since this formulation of the edit distance is defined for trees alone, we cannot apply it to FSMs and thus make a comparison. Using the fact that trees are a special type of graphs, we propose instead to use the Graph Edit Distance (GED). GED is defined as the minimum number of edit operations (add/remove/substitute nodes and edges) to execute on a graph $g_1$ to make it isomorphic to another graph $g_2$~\cite{abu-aisheh_exact_2015}. Using the standard tuple definition of graphs, $G = (V, E)$ with $V,E$ the sets of vertices and edges respectively, the GED between $g_1 = (V_1, E_1)$ and  $g_2 = (V_2, E_2)$, is defined as:
\begin{equation}
    GED(g_1, g_2) = \min_{e_1,\dots,e_k\in \gamma(g_1, g_2)} \sum_{i=1}^k c(e_i)
\end{equation}

Where $c$ is the cost of the edit operation $e_i$ and $\gamma(g_1, g_2)$ denotes the set of edit paths to transform a graph into the other. For this analysis, we use this measure as implemented in the \texttt{NetworkX} python library~\cite{hagberg_exploring_2008}.

\paragraph{Edit Distance in BTs}

Considering BTs as a graph, adding or removing one node $n^{\star}$ contributes with the editing of two elements in the graph, a vertex corresponding to the node and an edge from the node to the parent.
For the examples we brought, we have the following:
\begin{itemize}
    \item the BT in Fig.~\ref{fig:base_BT} has 14 nodes and 13 edges;
    \item the BT in Fig.~\ref{fig:tuck_BT} has 17 nodes and 16 edges;
    \item the BT in Fig.~\ref{fig:safer_BT} has 15 nodes and 14 edges;
    \item the BTs in Fig.~\ref{fig:dock_BT} and~\ref{fig:battery_BT} have 18 nodes and 17 edges;
\end{itemize}

Since we add 3 nodes for Fig.~\ref{fig:tuck_BT}, 1 for Fig.~\ref{fig:safer_BT}, and 4 for Fig.~\ref{fig:dock_BT} and~\ref{fig:battery_BT}, we obtain an edit distance of 6, 2, and 8, as reported in Table~\ref{tab:graph_edit_dist}. 

\paragraph{Edit Distance in FSMs}

In the case of an FSM, every new state $n^{\star}$ contributes with one vertex and at least 5 edges (4 for the alternative state case). If however we aim to edit a fully connected state, then the GED depends on the number of the states in the FSM. In such a case, the GED is $4+n$ (the state $n^{\star}$, the \textit{RUNNING} transition, transitions to and from SELECTOR, and a transition from every other state to $n^{\star}$).

Similarly to what done with BTs, we use the measure to compare FSMs of Fig.~\ref{fig:add_SM} with the baseline of Fig.~\ref{fig:fault_SM}. Such FSMs are directed graphs with the following elements:
\begin{itemize}
    \item the FSM in Fig.~\ref{fig:fault_SM} has 6 nodes and 18 edges;
    \item the FSMs in Fig.~\ref{fig:tuck_SM} and~\ref{fig:dock_SM} have 7 nodes and 22 edges;
    \item the FSMs in Fig.~\ref{fig:safer_SM} has 7 nodes and 21 edges;
    \item the FSM in Fig.~\ref{fig:battery_SM} has 7 nodes and 25 edges.
\end{itemize}

Where the outcome \textit{SUCCESS} is also considered as a node. We observe that adding a state in some intermediate point in the execution sequence (Fig.~\ref{fig:tuck_SM}) has no substantial differences as compared to adding it as a final state (Fig.~\ref{fig:dock_SM})
The values for the edit distances are reported in Table~\ref{tab:graph_edit_dist}.
\par

\paragraph{Edit Distance in HFSMs}
We make a similar but briefer analysis on HFSMs. With the example portrayed in Fig.~\ref{fig:HFSM}, we obtained the graph representation by converting every node as it follows:
\begin{itemize}
    \item an action node is a subgraph with 1 vertex and 3 edges. The vertex being the node itself and the edges being the 3 transitions, one from each return status to the next child or the parent return statuses;
    \item a condition node is similar to an action node, with one edge less because conditions do not return \textit{RUNNING};
    \item a control node contributes with 1 vertex and 4 edges. The additional edge is the transition from the node to the first child.
    \item we add 3 more states to account for the outcome of the execution, one for each return status.
\end{itemize}
We can thus compute the GED between two HFSMs as a function of the difference between their structural components:
\begin{equation}
    GED(hfsm_1, hfsm_2) = 3\Delta c + 4\Delta a + 5\Delta i,
\end{equation}
where $c$ are condition nodes, $a$ are action nodes, and $i$ are internal or control flow nodes.

With such design, we disregard the IN elements and the internal return statuses. Taking as example the HFSM that we would obtain from the BT in Fig.~\ref{fig:base_BT}, we would have:
\begin{itemize}
    \item 4 \textit{Fallback} nodes, contributing 1 vertex and 4 edges each, for a total of 4 vertices and 16 edges;
    \item 2 \textit{Sequence} nodes, contributing 2 vertices and 4 edges each, for a total of 2 vertices and 8 edges;
    \item 4 \textit{Action} nodes, contributing 1 vertex and 3 edges each, for a total of 4 vertices and 12 edges.;
    \item 4 \textit{Condition} nodes, contributing 1 vertex and 2 edges each, for a total of 4 vertices and 8 edges;
\end{itemize}
This makes a total of 17 vertices and 44 edges. To resume:
\begin{itemize}
    \item the HFSM representing the BT in Fig.~\ref{fig:base_BT} has 17 vertices and 44 edges;
    \item the HFSM representing the BT in Fig.~\ref{fig:tuck_BT} has 20 vertices and 53 edges;
    \item the HFSM representing the BT in Fig.~\ref{fig:safer_BT} has 18 vertices and 47 edges;
    \item the HFSM representing the BTs in Fig.~\ref{fig:dock_BT} and~\ref{fig:battery_BT} have 21 vertices and 57 edges;
\end{itemize}
The minimum GED for the considered examples is 4 when we add the behavior to move more safely, to a maximum of 17 when we add the subtrees for docking or recharging the batteries, as reported in detail in Table~\ref{tab:graph_edit_dist}.

\begin{figure*}[htbp]
     \centering
\begin{subfigure}[b]{\textwidth}
    \centering
    \includegraphics[width=.5\linewidth]{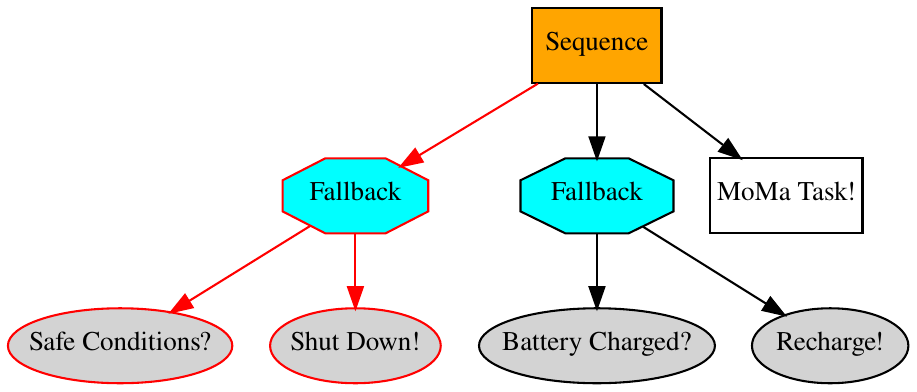}
    \caption{Adding the a high priority subtree to a BT does not affect the other behaviors in the policy.}
    \label{fig:priority_bt}
\end{subfigure}
\hfill
\begin{subfigure}[b]{\textwidth}
    \centering
    \includegraphics[width=.9\linewidth]{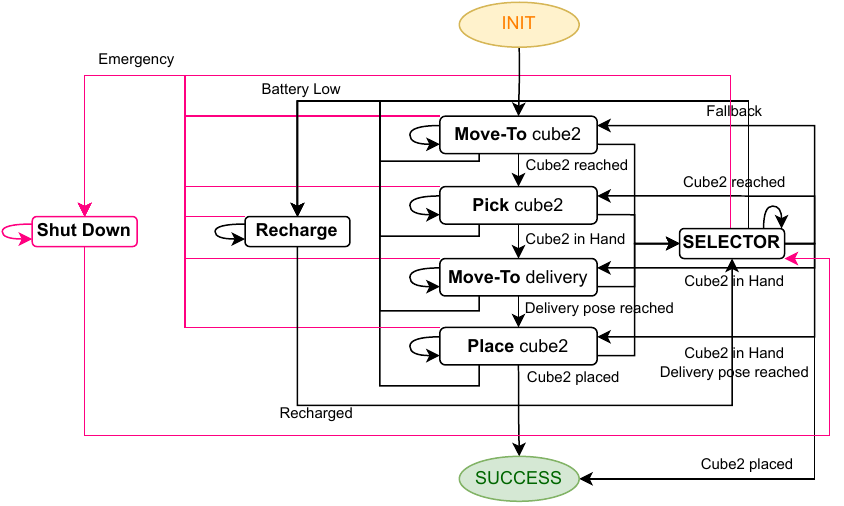}
    \caption{Adding connected states to the FSM complexifies the structures and the maintenability of the code.}
    \label{fig:priority_fsm}
\end{subfigure}
\caption{Editing high priority behaviors in a policy representation: modularity allows to interchange the priority of subtrees in a BT without requiring to modify other behaviors in the policy.}
        \label{fig:priority}
\end{figure*}

\vspace{.5cm}
To conclude, in Fig.\ref{fig:addall_BT} and~\ref{fig:addall_SM} we show the BT and FSM resulting from adding all the nodes described in the previous examples, respectively. It can be seen that the number of edits required to perform in an FSM increases with the number of states added and it soon becomes cumbersome to maintain, while this maintenance cost is mitigated in the case of a BT.

Another example that highlights the benefits of modularity in a BT, is when we need to add behaviors with high priority and later decide the priority order among them. In particular, we consider as a baseline the BT in Fig.~\ref{fig:battery_BT} and the FSM in Fig.~\ref{fig:battery_SM}. We now want the robot to shut down in case of emergency. We obtain the policies in Fig.~\ref{fig:priority} where in the BT the subtree realizing the mobile manipulation task of Fig.~\ref{fig:base_BT} is collapsed into anode called \textit{`Moma Task!'} for a better visualization and to stress that editing a BT does not modify other behaviors in the tree. Since a BT executes with tick signals that propagate from left to right, the subtree with the highest priority is placed in the leftmost part of the tree. To change the priority order it is enough to re-arrange the subtrees in the list of the children of the root node. In the case of the FSM, the priority depends on how the transitions are designed. If we want to change the priority order in the FSM of Fig.~\ref{fig:priority_fsm}, we would need to edit every implementation of the states to trigger the transition to `Recharge!' before the one to `Shut Down!', then to make a transition from `Shut Down!' to `Recharge!'. From a code design perspective, having copies of the same instruction block in multiple places is a source of error.

\subsection{Reactivity}

A reactive control policy should achieve two behaviors: 
\begin{enumerate}
    \item if during the task execution, part of the task is done or undone by an external agent or disturbance, the robot should react to it, either by jumping over the already executed steps, or by redoing those steps that have been undone;
    \item if during task execution, an action with a higher priority needs to be executed (i.e. recharge the robot if it's low on battery), the robot should interrupt the action that is currently executing to run the one with higher priority instead.
\end{enumerate}

Taking as example the BT in Fig.~\ref{fig:battery_BT}, where a subtree for handling battery recharge has been added (additional elements highlighted in red), it achieves both behaviors:
\begin{enumerate}
    \item if the place pose is in the neighborhood of the grasping pose, then the robot wouldn't need to execute the second navigation action as its condition would already be satisfied, thus jumping directly to the execution of the place action. The BT is ticked also if the root returns \emph{Success}, so if the robot completes the task but then an agent moves the cube from its target pose, then the robot would attempt grasping and then placing again.
    \item subtrees with high priority actions are found in the top left part of the tree. Since a BT is recursively ticked from left to right and nodes return \emph{Running}, any action that the robot is executing can be preempted as soon as the condition monitoring the battery status returns \emph{Failure}, thus transferring the execution to the `Recharge!' action.
\end{enumerate}

\begin{figure}[tbp]
    \centering
    \includegraphics[width=.9\linewidth]{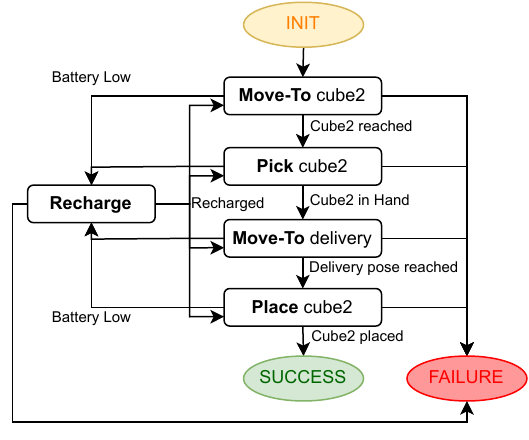}
    \caption{Non reactive FSM with $M_s = 4$ and $M_{fc} = 1$.}
    \label{fig:nonfault_SM}
\end{figure}

\begin{figure}[tbp]
    \centering
    \includegraphics[width=\linewidth]{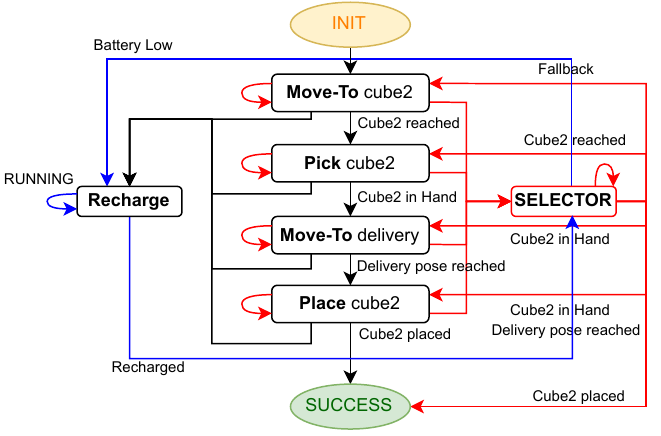}
    \caption{Reactive version of the FSM in Fig.~\ref{fig:nonfault_SM}. This is the same FSM of Fig.~\ref{fig:battery_SM}, were we highlight different elements. The red and blue elements correspond to the operations related to $M_s$ and $M_{fc}$ in Equation~\ref{eq:effort}, respectively.}
    \label{fig:effort_SM}
\end{figure}

In an FSM, the only way to achieve the same level of reactivity, is to have a transition from any state to every other, or to adopt a solution like the one proposed in Section~\ref{sec:design}. This is manageable for small problems, but it quickly grows out of hand if the task is complex. Moreover, with such a choice, modularity is sacrificed.
To quantify the price for reactivity, we can count the number of operations required to transform the sequential FSM of Fig.~\ref{fig:seq_SM} into the reactive version of Fig.~\ref{fig:fault_SM}. The former has 5 nodes and 4 edges, while the latter 6 nodes and 18 edges. The price for the transformation is a total of 15 operations (obtained by using the Graph Edit Distance measure). We can define this measure as \textit{Effort} and make it a function of the number of states. To transform an FSM with $M_s$ sequential states and $M_{fc}$ fully connected states in a reactive FSM, with our proposed design, the effort is:
\begin{equation}\label{eq:effort}
    E(M_s, M_{fc}) = 3(M_s + 1) + M_{fc}[(M_s + M_{fc} - 1) + 3].
\end{equation}
Let's consider the sequential FSM example of Fig.~\ref{fig:nonfault_SM} with $M_s = 4$ and $M_{fc} = 1$. Here, the fully connected state is the `Recharge' state. The effort to transform this FSM into the fault-tolerant version of Fig.~\ref{fig:effort_SM} requires the following operations:
\begin{enumerate}
    \item add the SELECTOR state plus a \emph{`RUNNING'} self-transition and a transition to the \emph{SUCCESS} outcome\----$3$ operations;
    \item for each $M_s$: add a \textit{RUNNING} self-transition, a \textit{FAILURE} transition from the state to SELECTOR and another one from SELECTOR back to the state to resume the execution\----$3$ operations;
    \item for each $M_{fc}$: remove transitions from this state to any other state\----$(M_s + M_{fc} - 1)$ operations\----then add the interaction to SELECTOR as in the previous point\----$3$ operations.
\end{enumerate}
Defining $M=M_s + M_{fc}$, Equation~\ref{eq:effort} can be rewritten as:
\begin{equation}
    E(M, M_{fc}) = 3(M + 1) + M_{fc}(M - 1).
\end{equation} 

To conclude, BTs are naturally reactive, so $E = 0$, independently on the number of nodes. The same holds for HFSM, that are reactive by construction as we enforced a structure.

\subsection{Readability}

To compare a BT with an FSM in terms of readability, we propose to compute the number of graphical elements in the two structures as a function of the robot skills (action behaviors in a BT and states in an FSM). A backchained BT (like the one in Fig.~\ref{fig:base_BT}) would usually have as many action behaviors as conditions (unless an action has two or more post-conditions or the same post-condition is achieved by more than one action, as in the case of Fig.~\ref{fig:safer_BT}). Then, since every post-condition is coupled with an action achieving that, we would have a \emph{Fallback} node to make each coupling. Finally, the number of \emph{Sequence} nodes is variable, as it depends on the number of actions that require expansion (i.e., those actions that have pre-conditions), but we can assume as a rule of thumb, that the number of \emph{Sequence} nodes is one half of the action behaviors. Thus, if we express the BT design complexity for $M$ action nodes, we would have a total of $N = 3.5M$ nodes. Those will count for \textit{Active Elements} in Table~\ref{tab:properties} (i.e., those elements that can be manipulated). If we had to consider all graphical elements, we should include every edge connecting each parent node to its children, for a total of $T = N-1$ connections. This makes it a total of $S = N + T = 7M-1$ elements.
\par
Using the same analysis for an FSM, for every state we would have a transition connecting it to the next one, with the last transitioning to a final \emph{Success} state, or the outcome of the task. To achieve full reactivity, we would need a \emph{Running} self-transition for each state. An additional SELECTOR state is also needed to allow the robot to reactively restart the execution where it stopped in case of errors, with a transition to this state from every other state. With such a design, the FSM described by $M$ action states, would have a total of $N = M + 1$ nodes (we add the SELECTOR state). Then we have the following transitions:
\begin{itemize}
    \item $N$ \textit{Running} transitions, one per state;
    \item $N - 1 = M$ \textit{Failure} transitions, one per action state;
    \item $N - 1 = M$ transitions from the SELECTOR state to any other action state to resume execution;
    \item $M$ transitions from every action state to the next;
    \item $1$ more transitions connecting the SELECTOR state to the outcome.
    \item $M-1$ more transitions for every fully connected state $M_{fc}$, thus amounting for $T_{fc} = M_{fc}(M-1)$
\end{itemize}
This makes a total of $T + T_{fc}$ transitions where $T = N + 3M + 1 = 4M + 2$ and a total of $S = T + N + 1 + T_{fc} = 5M + 4 + T_{fc}$ elements, where we also added the outcome state. 
An alternative design representing a reactive FSM as a fully connected graph, would have instead $N-1$ transitions for every state, making it a total of $S = M(M-1)$ elements.
\par
Since a HFSM is built from a BT, we will have one condition and one fallback node for every action and approximately half of it for sequence nodes. This means $M$ actions, $M$ conditions, $M$ fallback nodes and $0.5M$ sequence nodes. To compute the \textit{Active Elements} in a HFSM, we can use the same analysis we described to build the graph, in Section~\ref{sec:modularity}, where actions contribute for 10 elements, conditions for 7, control nodes for 8. For the \textit{Graphical Elements} we need to add 2 elements to each node to take into account the IN state plus a transition. Finally, for both cases we remove 3 transitions from the root return statuses. We thus obtain the numbers reported in Table~\ref{tab:properties}.
\par
The readability of a decision structure is often determined by how easy it is, for a human operator, to understand and debug the agent behavior just by looking at the structure. This definition is hard to quantify and it is often subjective. However, we will describe how an operator would behave if they had to supervise an agent controlled both by an FSM and a BT. Since for every state of an FSM, the possible next robot actions are univocally determined by the transitions from that state to the others, the operator has a clear understanding of what is going to happen. This is due to the fact that the execution in an FSM is causal (i.e., the execution of a state depends on the past events). With a BT instead, because of reactivity and the fact that at every iteration the execution resumes from the root, any action can be preempted and the execution flow can jump to any other part in the tree. To adapt to this behavior of the decision structure, an operator would certainly require more training. There are however functionalities that allow highlighting the currently executing behavior, together with its return status, to greatly support human supervision. As an example, with the modified version of \texttt{py\_trees} provided by authors in~\cite{iovino_programming_2022}\footnote{The code repository found under~\url{https://github.com/ethz-asl/bt_fsm_comparison}.}, at runtime the currently running behavior is highlighted in yellow, successful ones in green, and failed ones in red. In this way, tracking the robot execution is easier.
\par
Finally, even if the preference of a tree structure over a graph structure is subjective, the reader must remember that in the examples provided here, the robot has to achieve a very simple mobile manipulation task, so the graph representation might lose clarity with increasing task complexity since the number of intersections among the transitions will increase. With reference to Section~\ref{sec:experiments} for an example, we highlight the fact that in a tree structure it is possible to collapse subtrees into single nodes (as in Fig.~\ref{fig:full_bt}) to remove detail complexity. This is not a feasible option in FSMs due to the many transitions (as seen in Fig.~\ref{fig:full_fsm}), unless states are organised in hierarchies but with the limitations that HFSMs introduce (as explained in Section~\ref{sec:work}). 


\begin{figure}[tbp]
    \centering
    \includegraphics[width=.9\linewidth]{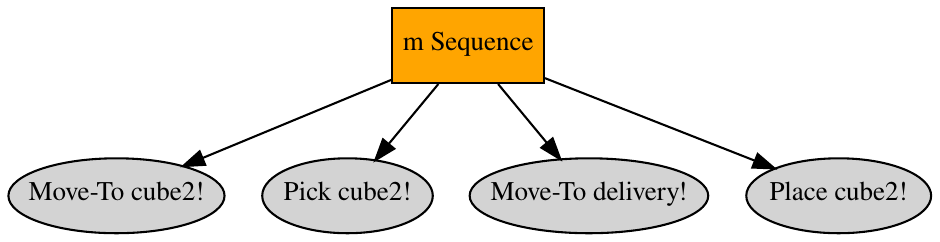}
    \caption{A BT rooted with a \emph{Memory Sequence} node, achieving open loop control.}
    \label{fig:seq_BT}
\end{figure}

\begin{figure}[tbp]
    \centering
    \includegraphics[width=.9\linewidth]{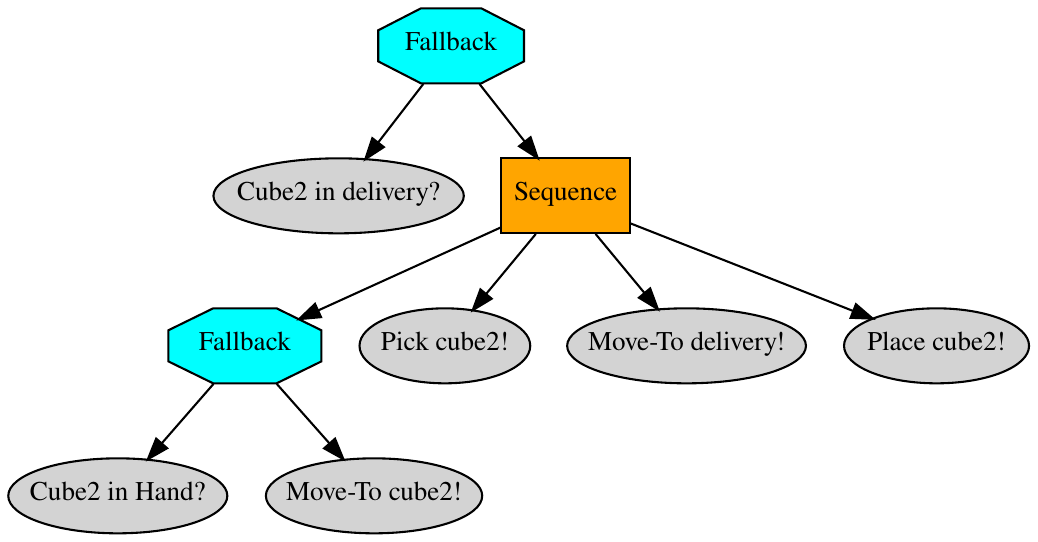}
    \caption{A BT that could be generated by Genetic Programming, according to the results in~\cite{iovino_learning_2021}.}
    \label{fig:gp_bt}
\end{figure}

\subsection{Comments on alternative designs}

FSMs have been around for much longer than BTs, becoming a standard in the design of control policies for robotics. Their success is mainly due to their design simplicity, which allows for intuitive understanding of the robot behavior. An FSM like the one in Fig.~\ref{fig:seq_SM} solves a mobile manipulation task sequentially, without granting fault tolerance or reactivity towards unforeseen events. Upon failure, the whole sequence has to be executed again. This design realizes a feed-forward control, as any feedback from the sensors is used as long as it triggers a transition, any unforeseen disturbance will not result in any transition, thus leading to a failure of the controller. If it is argued that simplicity and readability of the design are the most important aspects of a control policy, then the same feed-forward control can be achieved by a BT as well. By introducing the \emph{Memory} node (with symbol \emph{mSequence} or \emph{Sequence$^\star$}, instead of a \emph{Sequence} node) that remembers the state of the last executed child (thus not executing again a child that had previously succeeded), we would design a BT like the one in Fig.~\ref{fig:seq_BT}. If feed-forward control is desired, there is no benefit in using BTs over FSMs~\cite{colledanchise_behavior_2018}.
\par
Throughout this paper we have used a backchained BT design because of the associated convergence guarantees~\cite{ogren_convergence_2020}, but other designs may guarantee a more compact BT. One example is a BT generated by the Genetic Programming (GP) method proposed in~\cite{iovino_learning_2021} and used later in~\cite{styrud_combining_2022, iovino_framework_2023}. The learning method takes as input a task goal and a set of robot actions and conditions, then it evolves BTs to solve a mobile manipulation task similar to the one proposed as example for the analysis. If we were to use the same set of actions and conditions of this paper in the GP method, we would obtain a BT like the one in Fig.~\ref{fig:gp_bt}. We can observe that the GP learns to protect the action node driving the robot to the cube with a condition node (here, the condition checks that the cube is grasped), to avoid the chattering problem that we have previously discussed. This solution has the same properties of reactivity and modularity as the backchained one (as it is an inherent property of all BTs), but it is more compact and simpler as compared to its direct reactive FSM counterpart of Fig.~\ref{fig:fault_SM} (9 nodes and 8 edges versus 6 nodes and 18 edges).
\par
With that being said, it can be argued that a steeper learning curve is associated with using BTs. Thus, if the aim is to generate a control policy for a short task where the robot operates in a controlled environment, FSMs are likely to be easier to conceive and realize. If however the aim is to generate a robust control policy that might be subsequently modified or re-used, then BTs have a clear advantage. The design complexity can be mitigated by automatic generation methods that are already available in literature.

\section{Experiments and Results} \label{sec:experiments}

\begin{figure}[t]
    \centering
    \includegraphics[width=\linewidth]{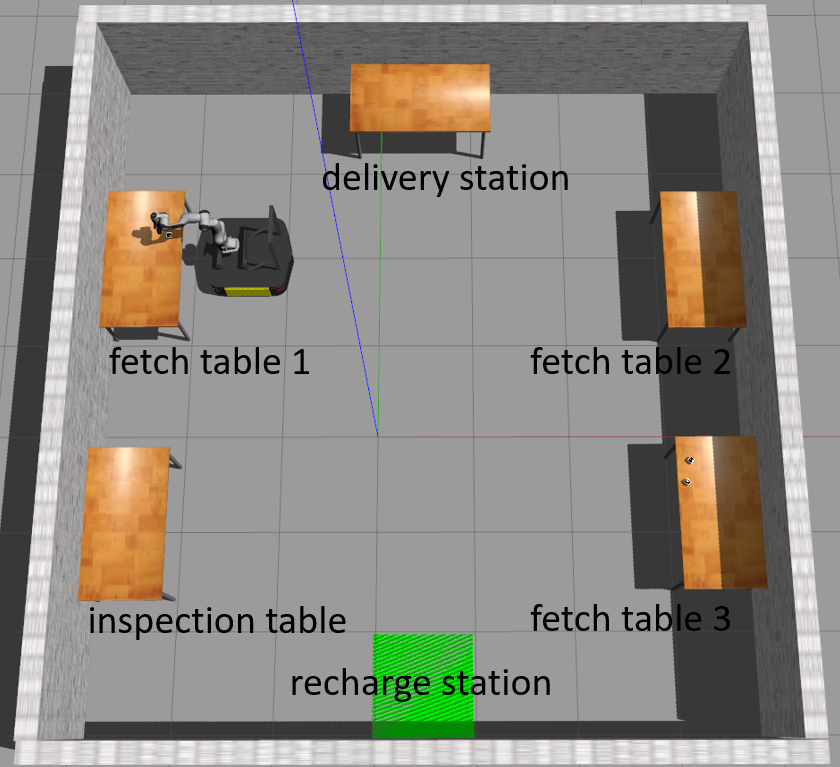}
    \caption{Simulation environment with the mobile manipulator implemented in the Gazebo simulator.}
    \label{fig:sim}
\end{figure}

In this Section, we conduct a comprehensive comparison between BTs and FSMs in the context of controlling a robot for a simulated mobile manipulation (MoMa) task by evaluating the metrics of Table~\ref{tab:properties} in concrete examples, with results summarized in Table~\ref{tab:experiments} . 
\par
We target tasks that are typical and representative of the MoMa domain.
Our primary focus is on the \textbf{Cleaning Up} task, a variant of the RoboCup@Home benchmark~\cite{wisspeintner_robocuphome_2009}, wherein the robot is tasked with collecting unknown objects scattered throughout a known environment. The complexity of the task requires the robot to navigate, recognize objects, manipulate them, and place them in a predefined area.
\par
We perform three sets of experiments, where the first and the second are taken from~\cite{iovino_programming_2022}. In the first set of experiments (Experiments~\ref{sec:exp1} to~\ref{sec:exp3}), we simplify the task to fetching a single known object, enabling a clear evaluation of both BTs and FSMs.
The evaluation methodology involves pairs of experiments, progressively increasing in complexity as new behaviors are incorporated into the policy. In this way, we wish to simulate a scenario where an engineer has to develop the policy representation for the task, starting from the MoMa baseline represented by the BT of Fig.~\ref{fig:base_BT} and the FSM of Fig.~\ref{fig:fault_SM}, and progressively modifying them.
To augment the task's complexity, two specific scenarios are introduced:
\begin{enumerate}
    \item \textbf{Battery Recharge Scenario:} The robot is required to recharge its batteries if they deplete at any point during task execution.
    \item \textbf{Docking Scenario:} Upon completing the task, the robot must dock at a specified location.
\end{enumerate}
The results of these experiments are supported by the accompanying video published with~\cite{iovino_programming_2022}\footnote{\url{https://www.youtube.com/watch?v=Z0GAkClVx-I}}.
\par
In the second set of experiments (Experiment~\ref{sec:exp4}) we focus on scalability: the effort required to modify the structures if they are already complex. Finally, in Experiment~\ref{sec:exp5} we expand experiments in~\cite{iovino_programming_2022} by transferring the simulated task on a real scenario. 
\par
The simulation environment (shown in Fig.~\ref{fig:sim}) features a robot comprising a Franka Emika Panda arm mounted on an omnidirectional Clearpath Ridgeback base. Equipped with an Intel RealSense RGBD camera, the arm's end-effector is instrumental in executing the manipulation tasks. In the simulation, the robot docks at the \textit{`inspection table'} and places the object at the \textit{`delivery station'}.

\begin{table*}[!ht]
\centering
\caption{Performances of the policies designed for the experiments according to the metrics defined in Table~\ref{tab:properties}. In the `Development' experiment, we gradually complexify the baseline structure simulating an engineer that progressively adds features to the base design (numbers in parenthesis quantify the steps such additions require). We first add a recharging behavior and then a docking one. In the `Scalability' experiment, we add a recharging behavior to a more complex baseline structure.}

\begin{tabular}{c c c c c c}
& & \multicolumn{4}{c}{\fontsize{10}{0}\selectfont \textbf{Metrics (BT/FSM)}}\\
\cmidrule{3-6}
\multicolumn{2}{c}{\fontsize{10}{0}\selectfont \textbf{Experiment}} &
\multicolumn{1}{c}{Cyclomatic Complexity (CC)} & 
\multicolumn{1}{c}{Edit Distance (ED)} & 
\multicolumn{1}{c}{Graphical Elements (+/-)} &
\multicolumn{1}{c}{Active Elements (+/-)} \\
\cmidrule(r){1-2} \cmidrule{3-6}
\multirow{3}{*}{Development} &
\multicolumn{1}{c}{Baseline} &
 1 / 14 & - / -&  27 / 24 & 14 / 24  \\
\cmidrule(lr){3-6}
& \multicolumn{1}{c}{Recharge} &
1 / 20 & 8 / 8 & 35 (+8) / 32 (+8) & 18 (+4) / 32 (+8) \\
\cmidrule(lr){3-6}
& \multicolumn{1}{c}{Docking} &
1 / 24 & 6 / 8 & 41 (+6) / 38 (+6) & 21 (+3) / 38 (+6) \\
\midrule
\multirow{2}{*}{Scalability} &
\multicolumn{1}{c}{Baseline} &
1 / 68 & - / - & 153 / 114 & 77 / 114 \\
\cmidrule(lr){3-6}
& \multicolumn{1}{c}{Recharge} &
1 / 92 & 6 / 26 & 159 (+6) / 140 (+26) & 80 (+3) / 140 (+26) \\
\bottomrule
\end{tabular}
\label{tab:experiments}
\end{table*}

\subsection{Skill Description}
As the paper's primary focus is on comparing the high-level behavior of control policies, certain assumptions are made to streamline the execution of low-level skills. In terms of navigation, the robot is constrained to a specific set of target poses, with each pose corresponding to a station or table within the environment of Fig.~\ref{fig:sim}. The navigation functionality is realized through the utilization of the ROS Navigation Stack.
AprilTags are employed to locate items within the environment. The trajectory planning for manipulation is implemented using ROS MoveIt!. To compute the grasping points, a shape-based grasp synthesis method named the Volumetric Grasping Network (VGN)~\cite{breyer2020volumetric} is employed.

Within the ROS framework, skills are instantiated as action servers. A client interface enables the interaction between the policy (BT or FSM) and the low level implementation of the skills in terms of initialization, goal transmission, status monitoring during execution, and goal cancellation.

We define the action skills and stauts checks as it follows:
\paragraph{\textbf{Move-To!}} this action allows the robot to navigate to a target pose given explicitly or reconstructed from the object marker.  

\paragraph{\textbf{Robot-At?}} this condition checks that the robot reached the desired pose with the desired $[x, y, yaw]$ tolerance.

\paragraph{\textbf{Pick!}} this action moves the robot arm to a target pre-grasp pose on top of the target object. Then, VGN is used to compute the grasping pose that is sent to the robot arm. Once the grasp is successful, the arm is tucked in a compact configuration for navigation.

\paragraph{\textbf{In Hand?}} this condition checks if the robot is holding an object.

\paragraph{\textbf{Place!}} this action moves the robot arm to a target pose and drops the object. Then the robot moves the arm to a configuration that allows it to monitor the scene and evaluate the end pose of the object with the wrist camera.

\paragraph{\textbf{Object-At?}} this condition checks that the object is at the desired pose with the desired $[x, y, z]$ tolerance.

\paragraph{\textbf{Recharge!}} this action makes the robot navigate to the recharge station and then it instantaneously fills up the battery level.

\paragraph{\textbf{Battery Lv?}} this condition determines when the robot shall recharge its batteries.

\paragraph{\textbf{Dock!}} this action makes the robot dock at the \emph{`inspection table'}. The achievement of the action is monitored by `Robot-At?'.

For the Scalability experiment, we also include the following:
\paragraph{\textbf{Search!}} this action makes the robot follow 5 viewpoints (one per table in Fig.~\ref{fig:sim}), where the robot stops and inspects the table with the camera to look for markers.

\paragraph{\textbf{Found?}} this condition determines which markers to search for and by consequence to stop the search behavior once they are all found.

\subsection{Experiment 1: Baseline} \label{sec:exp1}

In this experimental scenario, the robot is tasked with retrieving a cube located on \emph{`fetch table 1'} as illustrated in Fig.~\ref{fig:sim}, and transporting it to the designated \emph{`delivery station'}. The robot initiates the task from the room's center, and the cube's predetermined position is known in advance. This specific task is also attempted using the sequential FSM depicted in Fig.~\ref{fig:seq_SM}. The BT and the fault-tolerant FSM employed for solving the task are presented in Fig.~\ref{fig:base_BT} and Fig.~\ref{fig:fault_SM}, respectively.

The sequential FSM successfully accomplishes the task only under the condition that all actions execute correctly and within defined tolerances. Failures during execution necessitate a task reset. In contrast, the BT and the fault-tolerant FSM reattempt failed actions. Additionally, due to the recursive ticking from the root in the BT, it continues to execute even upon successful completion. This characteristic enables the robot to respond to unexpected changes, such as a human operator relocating the cube to another table after task completion, allowing the robot to adapt and retrieve the cube again, provided it knows its updated location.

In terms of computational complexity when adding behaviors to the base task, the BT in Fig.~\ref{fig:base_BT} consists of a graph with 14 nodes and 13 edges, while the FSM in Fig.~\ref{fig:fault_SM} has 6 nodes and 18 edges, with a cyclomatic complexity of 14.

\subsection{Experiment 2: Recharge Scenario} \label{sec:exp2}
In this experiment, a recharging behavior is introduced to the existing manipulation task, necessitating the engineer to modify the policy as depicted in Fig.~\ref{fig:battery_BT} and Fig.~\ref{fig:battery_SM}. In the case of the BT, the engineer needs to insert the recharge subtree into a new \emph{Sequence} root node, prioritizing the recharging subtree over the subtree handling the manipulation task. This involves eight elementary operations: creating the four nodes (root node and nodes in the left subtree) in Fig.~\ref{fig:battery_BT}, adding two leaves to the \emph{Fallback} node, and finally incorporating the recharge subtree and the pre-existing BT for the MoMa task into the new root.
\par
It's important to note that the assumption is made that the recharge behavior instantaneously replenishes the batteries. This assumption could be justified by scenarios where an operator swiftly switches the robot's batteries when it halts at the recharge station. If the batteries are gradually recharged while the robot remains in the recharge station, the recharge subtree could be designed differently to preempt charging if a more critical task arises. An example of the BT implementation of this logic is proposed in~\cite{iovino2023learning}.
The BT in Fig.~\ref{fig:battery_BT} comprises 18 nodes and 17 edges, resulting in an ED of 8 with respect to the baseline.
\par
For the fault-tolerant FSM, eight operations are also required: creating the recharge state and the \textit{Running} transition, adding a transition from the recharge state to the SELECTOR state, and from every state to the recharge state. Two considerations emerge. Firstly, the number of new transitions depends on the number of existing nodes, unlike the BT where the operation count is independent of tree size. Secondly, the internal logic of the states must be adjusted to implement the case triggering a transition to the new state. This underscores a distinct advantage of BTs, where the switching logic is explicitly implemented in the representation.
The FSM in Fig.~\ref{fig:battery_SM} encompasses 7 nodes and 25 edges, resulting in an ED of 8 with respect to the baseline and has a cyclomatic complexity of 20.
\par
From a behavioral standpoint, the task execution remains unchanged, and the robot successfully proceeds to recharge the batteries if their level falls below $20\%$.

\begin{figure*}[tbp]
     \centering
\begin{subfigure}[b]{0.5\textwidth}
    \centering
    \includegraphics[width=\linewidth]{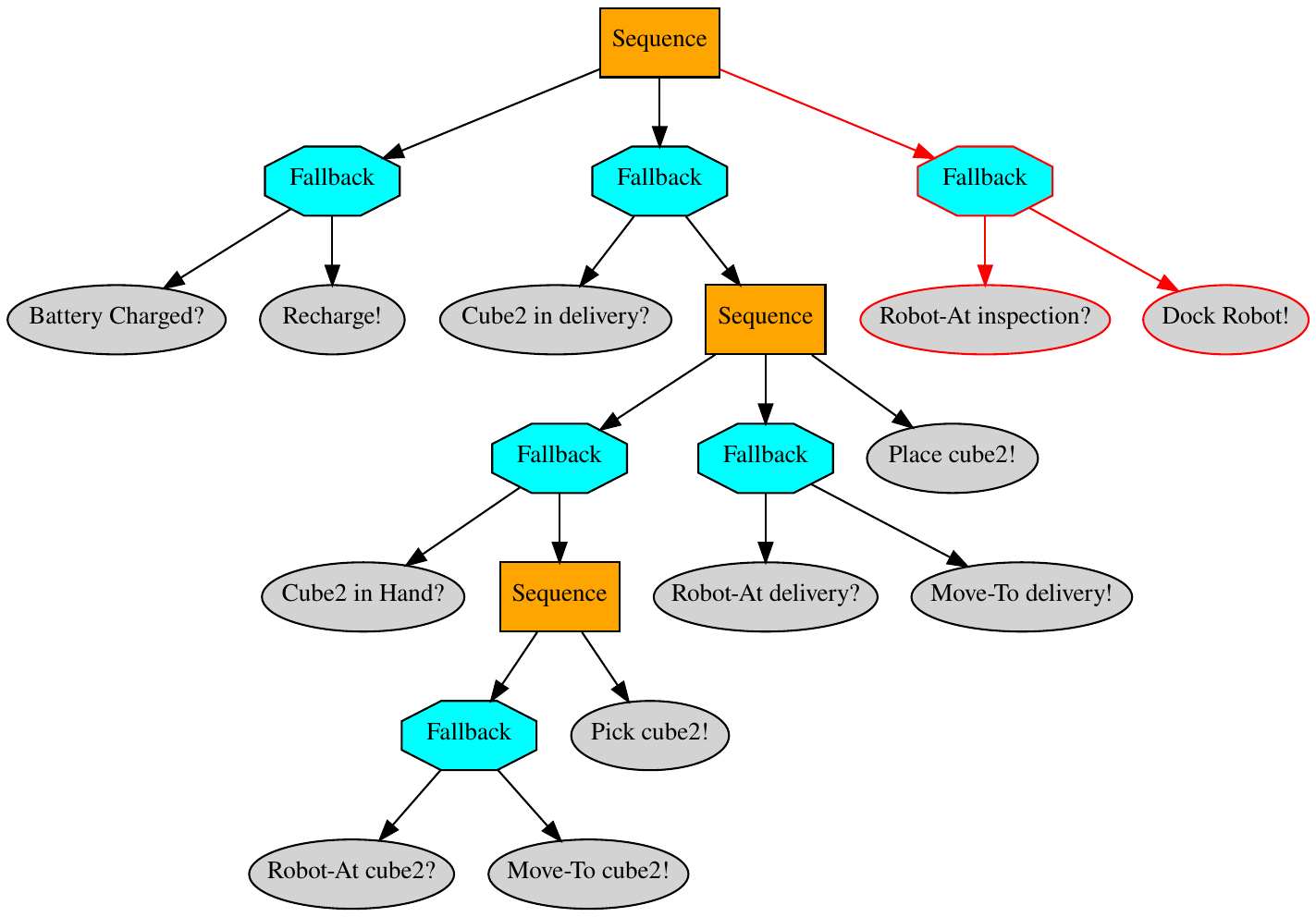}
    \label{fig:exp_BT}
\end{subfigure}
\hfill
\begin{subfigure}[b]{0.45\textwidth}
    \centering
    \includegraphics[width=\linewidth]{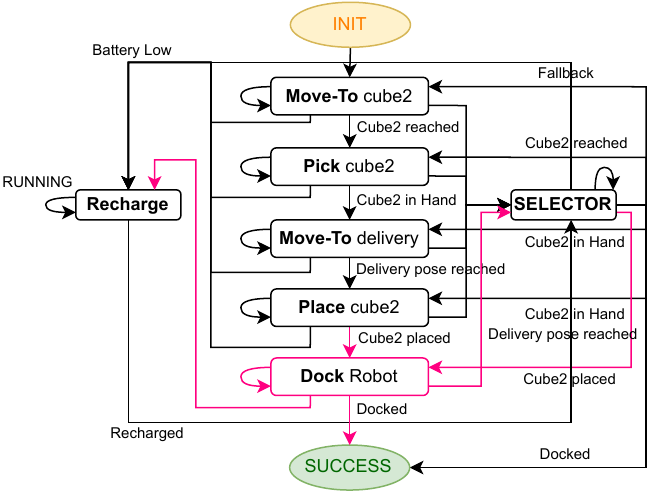}
    \label{fig:exp_SM}
\end{subfigure}
\caption{BT and FSM for the Experiment 3 obtained by adding a docking behavior to the representation of Experiment 2.}
\label{fig:exp3}
\end{figure*}

\subsection{Experiment 3: Docking Scenario} \label{sec:exp3}
In this instance, we introduce a docking behavior, as illustrated in Fig.~\ref{fig:exp3}. Since the \emph{Sequence} root node already exists, editing the BT entails: (1) creating the \emph{Fallback} node, the condition, and the action, (2) adding the two leaves to the control node, and (3) appending the subtree to the root. Once again, the remainder of the tree remains unaffected by this addition.
\par
For the FSM, apart from creating the new state, the \emph{Running} transition, and the transitions to and from the SELECTOR state, it is necessary to eliminate the transition from the \emph{`Place'} state to the \emph{Success} outcome and create a transition from the \emph{`Dock'} state to the outcome.
\par
In this experiment, similarly to the preceding cases, there is no significant divergence in the executed robot behavior between the two policies. The BT consists of 21 nodes and 20 edges, resulting in an $ED=6$ compared to the previous case of Experiment~\ref{sec:exp2}. On the other hand, the FSM comprises 8 states and 30 transitions, yielding an $ED=6$ and $CC=24$.

\begin{figure*}[htbp]
     \centering
\begin{subfigure}[b]{\textwidth}
    \centering
    \includegraphics[width=\linewidth]{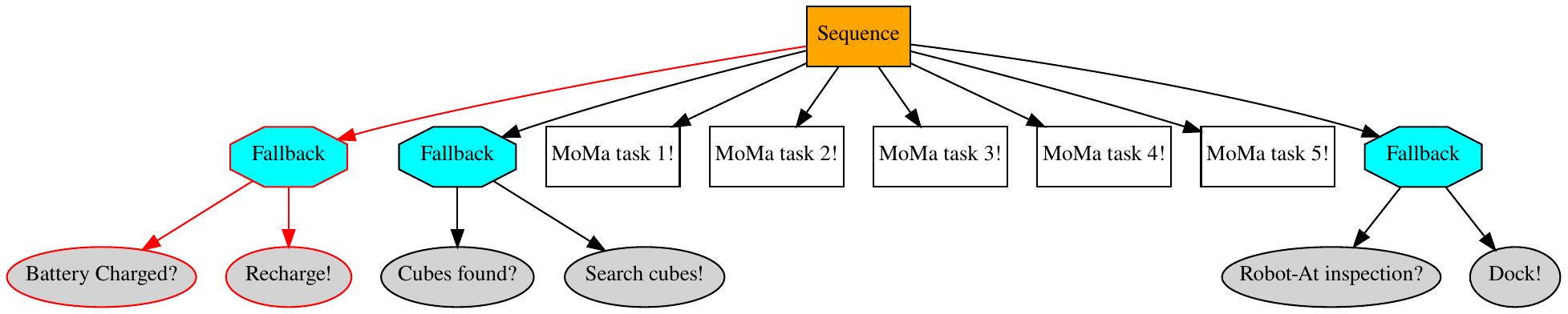}
    \caption{BT fetching five cubes, comprising the recharge and docking behaviors.}
    \label{fig:full_bt}
\end{subfigure}
\hfill
\begin{subfigure}[b]{\textwidth}
    \centering
    \includegraphics[width=\linewidth]{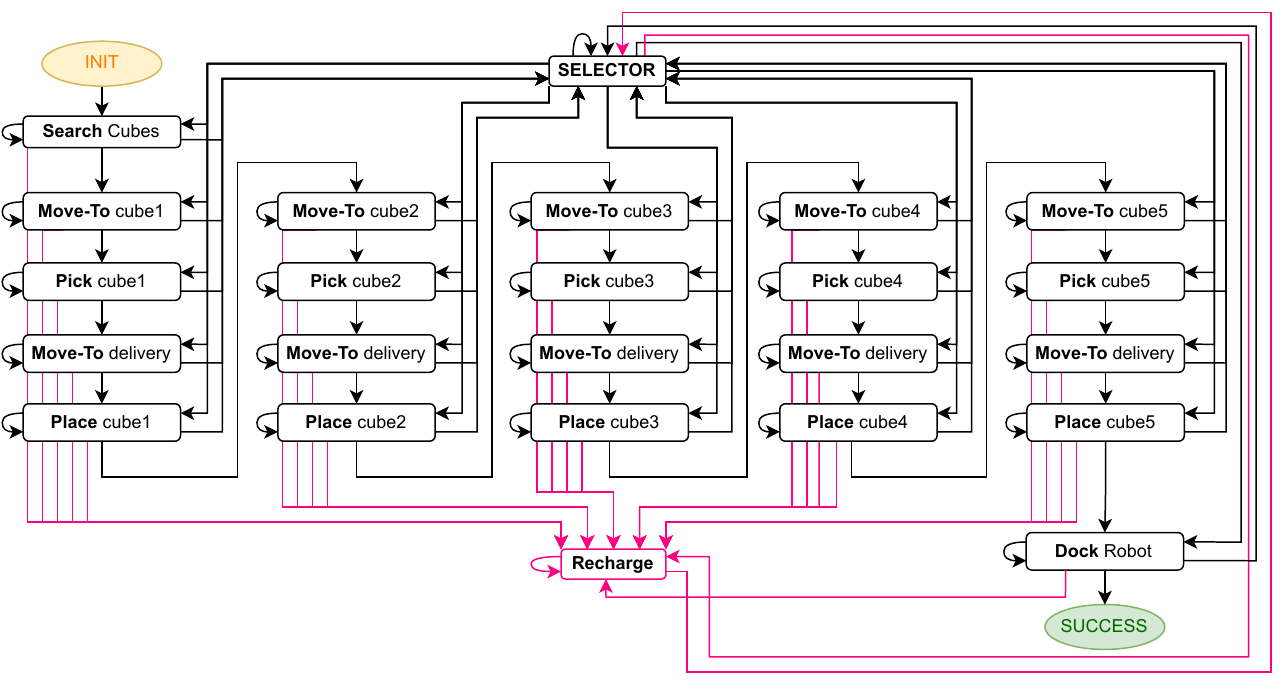}
    \caption{FSM fetching five cubes, comprising the recharge and docking behaviors. Transitions are omitted to improve readability.}
    \label{fig:full_fsm}
\end{subfigure}
\caption{\textbf{Experiment 4: Scalability.} In FSMs, adding connected states has complexity dependant on the number of states in the representation, resulting in hard to read structures. In BTs, this complexity is constant.}
\label{fig:scalability}
\end{figure*}

\subsection{Experiment 4: Scalability} \label{sec:exp4}
In the previous experiments, we deliberately focused on a scenario involving the retrieval of a single item. While this simplified task allowed for a detailed examination of implementation and programming efforts, it may not fully capture the advantages of modularity that BTs offer over FSMs. As we've noted, the insertion operations in FSMs are dependent on the existing number of states and transitions within the FSM.
\par
Expanding the task complexity to a scenario where the robot is required to search for and fetch five cubes before docking, we observe a BT with 77 nodes and 76 edges, and an FSM with 24 nodes and 90 transitions, considering the presented set of skills. At this point, incorporating a recharge behavior into the existing policy representations becomes more cumbersome for an FSM. This is because adding the transition to the recharge state from all other states is necessary. Consequently, the BT would have an $ED=6$ (similar to Experiment~\ref{sec:exp3}, where the root is already present), while the FSM would have an $ED=26$. This is due to the final BT (shown in Fig.~\ref{fig:full_bt}) featuring 80 nodes and 79 edges, and the final FSM  (shown in Fig.~\ref{fig:full_fsm}) comprising 25 nodes and 115 transitions.

\subsection{Experiment 5: Real Case Scenario} \label{sec:exp5}

\begin{figure}[htbp]
    \centering
    \includegraphics[width=\linewidth]{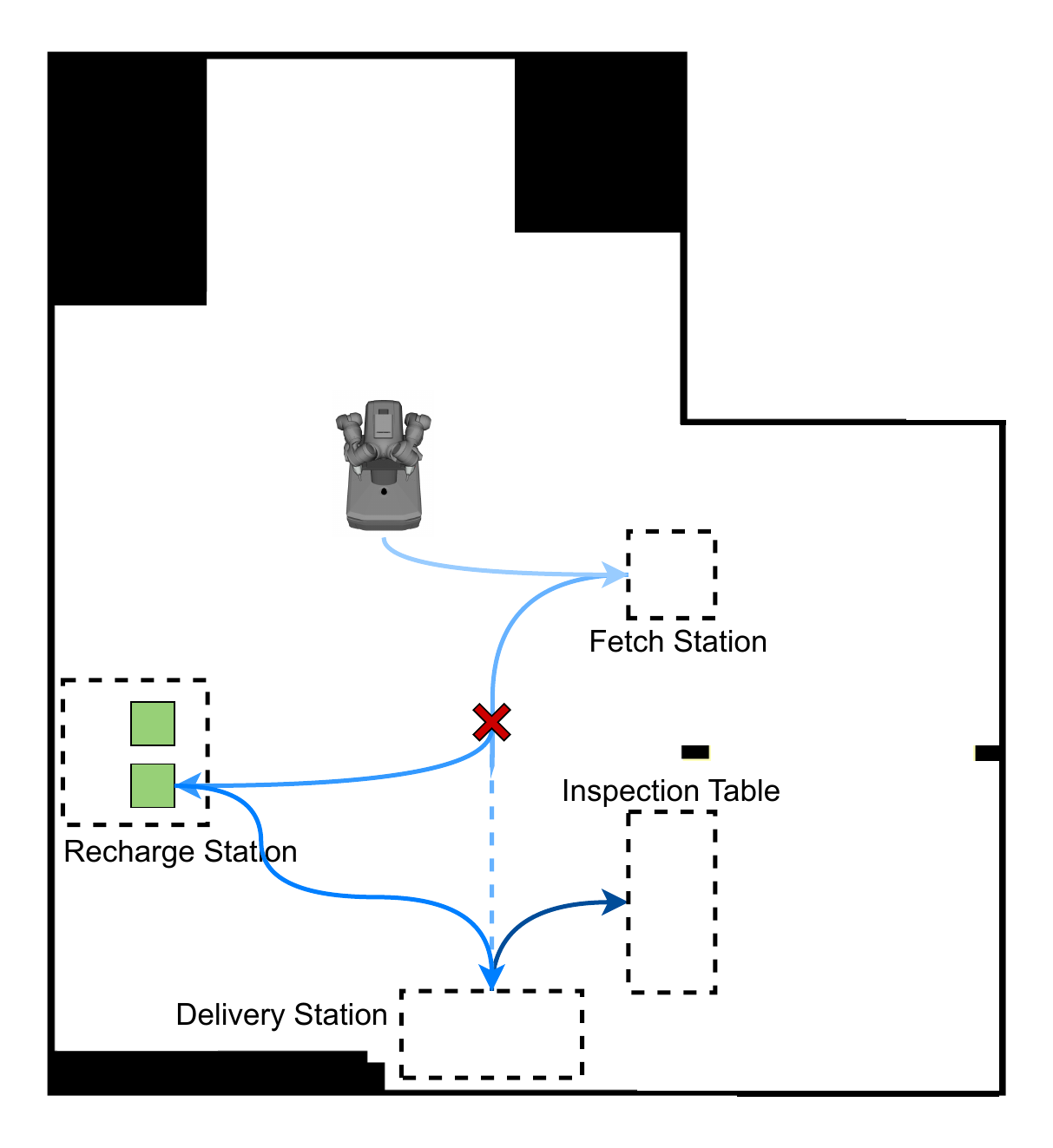}
    \caption{A schematic representation of the execution of Experiment~5 in the laboratory environment of the WARA Robotics. The solid black lines and areas represent walls and non accessible spaces and are therefore considered as obstacles by the robot. The arrows represent the trajectory followed by the robot during the task.}
    \label{fig:wara}
\end{figure}

\begin{figure*}[htbp]
    \centering
\begin{subfigure}[b]{.2\textwidth}
    \centering
    \includegraphics[width=\linewidth, height=4.7cm]{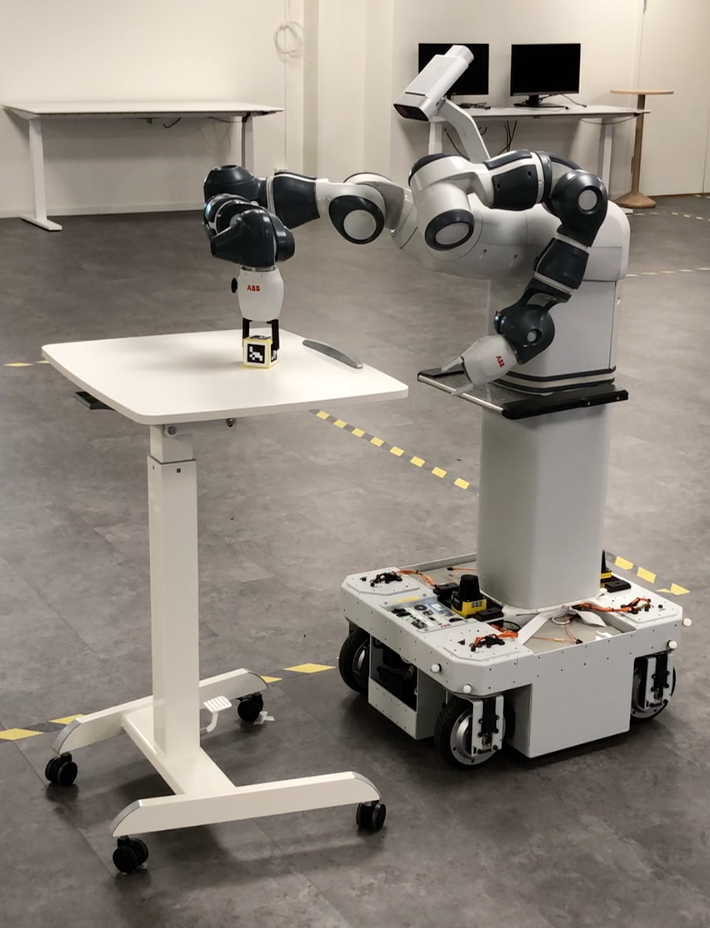}
    \caption{The robot is picking the cube at the fetch table.}
    \label{fig:myumi_pick}
\end{subfigure}
\quad
\begin{subfigure}[b]{.3\textwidth}
    \centering
    \includegraphics[width=\linewidth]{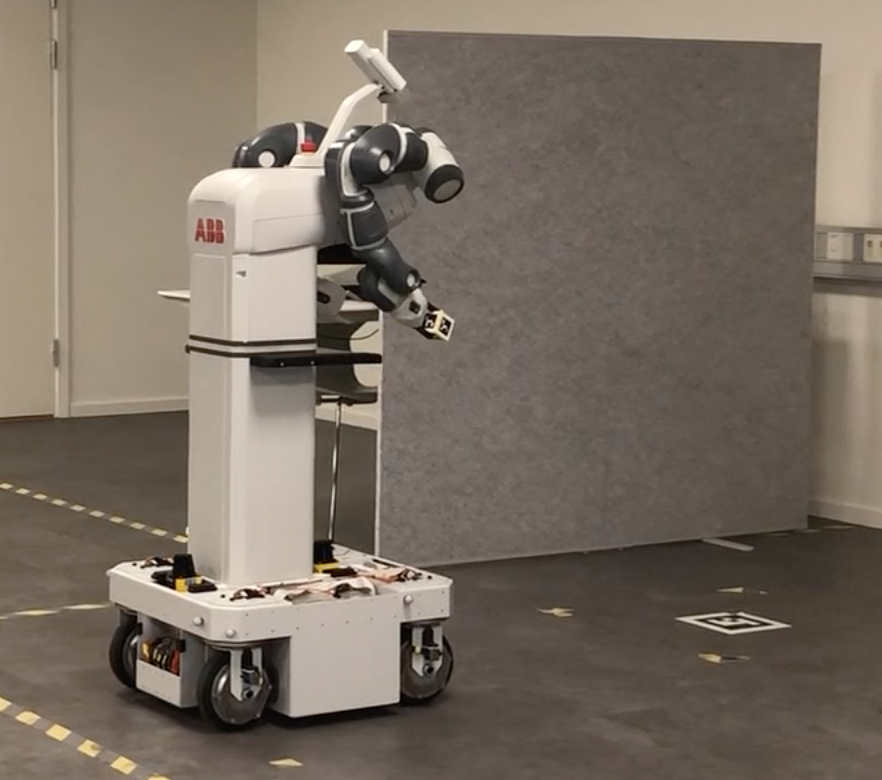}
    \caption{The robot is recharging the batteries at the recharge station.}
    \label{fig:myumi_recharge}
\end{subfigure}
\quad
\begin{subfigure}[b]{.2\textwidth}
    \centering
    \includegraphics[width=\linewidth, height=4.7cm]{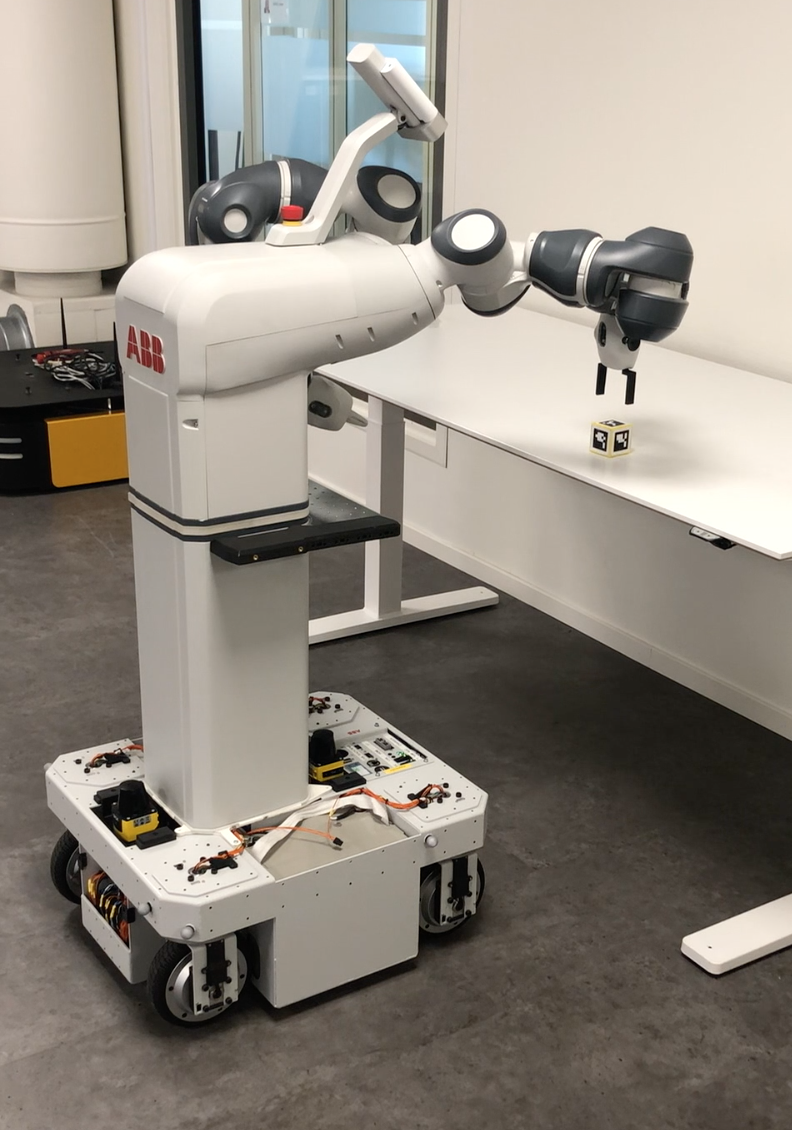}
    \caption{The robot is placing the cube at the delivery station.}
    \label{fig:myumi_place}
\end{subfigure}
\quad
\begin{subfigure}[b]{.22\textwidth}
    \centering
    \includegraphics[width=\linewidth, height=4.7cm]{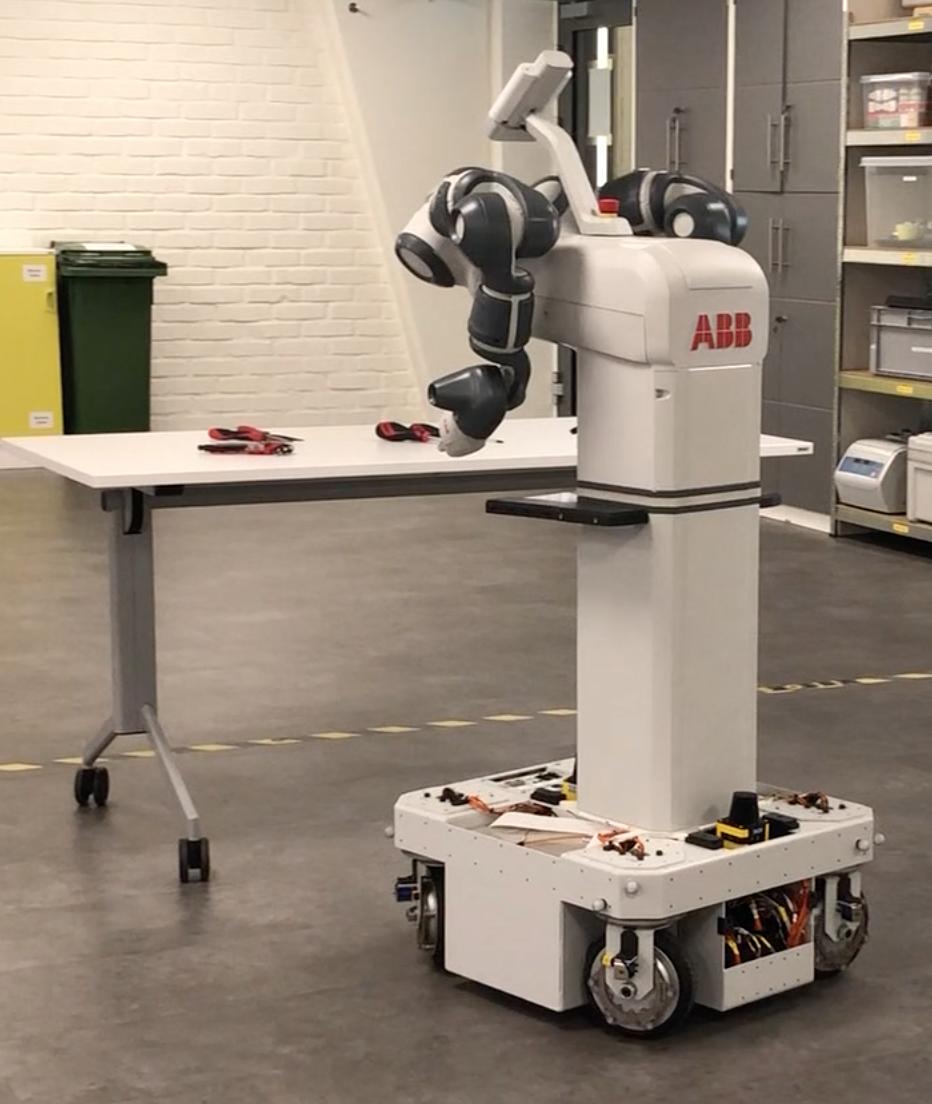}
    \caption{The robot is docking at the inspection table.}
    \label{fig:myumi_dock}
\end{subfigure}
\caption{\textbf{Experiment 5: Real Case Scenario.} For this experiment the ABB Mobile YuMi Research Platform is used. The Experiment is carried out in the environment of Fig.\ref{fig:wara}.}
\label{fig:myumi}
\end{figure*}

In this experiment we implement the same policies of Fig.\ref{fig:exp3} from Experiment~3 on a real robotic system. The goal of this experiment is to show the following:
\begin{enumerate}
    \item since the policies are a plan at the task level, their representation as a BT or an FSM are platform-agnostic. Only the low level implementation of the skills are platform-dependent.
    \item as shown in the previous experiments, provided that the implementation of the skills is the same, the robot behaves similarly if controlled by a BT or an FSM.
\end{enumerate}

To this extent, we run the experiment in the WASP Research Arena for Robotics (WARA Robotics\footnote{\url{https://wasp-sweden.org/industrial-cooperation/research-arenas/wara-robotics/}}, Fig.~\ref{fig:wara}) where we use an ABB Mobile YuMi Research Platform (shown in Fig.~\ref{fig:myumi}). The robot skills are implemented in the ROS2 framework. The Nav2 library is used for navigation, ArUco marker detection is used for the perception, and for manipulation proprietary ROS2 drivers are used to send pose targets to the robot controller that computes and executes a feasible trajectory and handles the opening and closing of the grippers.

\begin{algorithm2e}[!ht]
\caption{Pseudo-code of the MoveTo behavior in the BT.}\label{alg:moveBT}
\SetStartEndCondition{ }{}{}%
\SetKwIF{If}{ElseIf}{Else}{if}{:}{elif}{else:}{}%
\SetKwIF{IfNot}{ElseIfNot}{}{if not}{:}{else if not}{:}{}
\AlgoDontDisplayBlockMarkers\SetAlgoNoEnd\SetAlgoNoLine%
\SetKwFunction{init}{initialise}
\SetKwFunction{update}{update}
\SetKwFunction{move}{move}
\SetKwFunction{done}{navigation\_done()}
\SetKwFunction{success}{navigation\_succeeded}
\SetKwProg{define}{def}{:}{}
\define{\init{skill, goal, reference}}{
    $skill$.\move{goal, reference}
}
\BlankLine
\define{\update{skill, goal}}{
    \IfNot{$skill$.\done}{
        \KwRet RUNNING
    }
    \ElseIf{$skill$.\success{goal}}{
        \KwRet SUCCESS
    }
    \Else{
        \KwRet FAILURE
    }
}
\end{algorithm2e}

\begin{algorithm2e}[!ht]
\caption{Pseudo-code of the MoveTo state in the FSM.}\label{alg:moveFSM}
\SetStartEndCondition{ }{}{}%
\SetKwIF{If}{ElseIf}{Else}{if}{:}{elif}{else:}{}%
\SetKwIF{IfNot}{ElseIfNot}{}{if not}{:}{else if not}{:}{}
\SetKwComment{comm}{$\#\ $}{}
\AlgoDontDisplayBlockMarkers\SetAlgoNoEnd\SetAlgoNoLine%
\SetKwFunction{exec}{execute}
\SetKwFunction{battery}{battery\_low()}
\SetKwFunction{move}{move}
\SetKwFunction{done}{navigation\_done()}
\SetKwFunction{success}{navigation\_succeeded}
\SetKwProg{define}{def}{:}{}
\define{\exec{skill, goal, reference, recharge\_state}}{
    \comm{\small if low batteries trigger transition to recharge state}
    \IfNot{$skill$.\battery}{
         \KwRet   recharge\_state
    }
    \comm{\small send the goal only the first time}
    \IfNot{goal\_sent}{
        $skill$.\move{goal, reference}
    }
    \IfNot{$skill$.\done}{
        \KwRet RUNNING
    }
    \comm{\small if goal reached trigger transition to next state}
    \ElseIf{$skill$.\success{goal}}{
        \KwRet  SUCCESS
    }
    \Else{
        \KwRet FAILURE
    }
}
\end{algorithm2e}

To better understand the details of the implementation, we report the pseudo code of a behavior in the BT and a state in the FSM responsible to move the robot base to an input goal in Algorithms~\ref{alg:moveBT} and~\ref{alg:moveFSM} respectively. Note that, according to the \texttt{py\_trees} API for BTs, the \texttt{initialise} function is called only when the behavior is ticked the first time, then the \texttt{update} function is called for following ticks. In case of the FSM instead, the \texttt{execute} function is called cyclically until the execution transitions to another state. It can be appreciated how we designed a single API for all the skills to then wrap them in a behavior or state class for BT and FSM respectively. By looking at the proposed implementation, a behavior in a BT does not require any information about other behaviors. It is the BT structure through its control nodes that switches the execution to a child or another, for instance to the recharge behavior in case the batteries are low. In the case of the FSM instead, it is necessary to include a status check inside the state execution to trigger a transition to the recharge state. Moreover, this type of check has to be done for every other state in the FSM thus making it harder to maintain if its complexity increases. To add a state to the FSM, it is required to provide a the transitions mapping between the return statuses and other states in the FSM. In the case of Algorithm~\ref{alg:moveFSM}, \emph{SUCCESS} triggers a transition to the next state, while \emph{FAILURE} triggers a transition to the SELECTOR state.

As for Experiment~3, the robot has to fetch an item to then deliver it and finally dock. In Fig.~\ref{fig:wara} we show all these steps. The robot navigates to the \textit{`fetch station'} to pick up the target object. Then, it attempts to navigate to the \textit{`delivery station'} to place it but during the execution the batteries run low. The robot then interrupts current execution to prioritize the recharging. Once the batteries are loaded again, it resumes the task to first deliver the target object and finally dock at the \textit{`inspection table'}. Note that the depleting of the batteries is simulated and recharging them is instantaneous. This choice does not impact the argumentation for this experiment.
As in the case of the other experiments, the behavior of the robot does not change if it is controlled by a BT or an FSM. Some excerpts of the task execution are reported in Fig.\ref{fig:myumi}, while the full task is shown in the accompanying video, together with the task switching mechanism of BTs and FSMs.

\section{Conclusion} \label{sec:conclusion}

In this paper we compared Behavior Trees (BTs) and Finite State Machines (FSMs) in terms of the properties of modularity, reactivity, and readability with a set of examples drawn from robotic applications. We chose common metrics to quantify the differences between the two policy representations, derived upper bounds for the comparison, and highlighted the advantages of using BTs as an alternative to FSMs.
\par
Modularity is achieved when all single components of a software share the same common structure (e.g., the same type and number of inputs and outputs). Modularity is an important feature because it allows to reuse pieces of software with minor modifications. In a BT, a modular policy representation, single nodes as well as subtrees receive a tick signal as input and return a status signal. FSMs are not modular because every state switching is realized with transitions that are state-specific. Therefore, in a BT, any subtree can be independently modified, tested and formally verified with verification architectures like Linear Temporal Logic (LTL). On the other end, editing a state in a FSM requires to take care of all the related transitions.
\par
Reactivity is the ability to take the right action as a consequence of unexpected events or failed actions that might arise at run time. Reactivity allows a robot to be fault-tolerant. FSMs are not fault-tolerant by design, but can achieve fault-tolerant behaviors if for instance all the states are connected to each other. We proposed an alternative design for a fault-tolerant FSM and derived the effort required to modify the standard FSM to the alternative design as a function of the states. Then, we used this design to compare to BTs. With the proposed design FSMs behave like BTs, but BTs are inherently reactive.
\par
In a policy representation, readability is an important feature both at run time and offline. It allows human operators to understand what the robot is doing while solving a task or to analyse the policy to identify fallacies. FSM are more intuitive because of their sequential execution flow. However, as the task complexity increases and the structure grows in size, the high number of transitions jeopardize the tracking of the execution flow. BTs on the other hand, have a higher learning curve because it is necessary to understand the functioning of the control nodes to follow the behavior switching. However, the tree structure allows users to recognise the task hierarchy. 
\par
For the experimental part, we conducted the comparison using a set of variations of the \textbf{Cleaning Up} task as defined in the RoboCup@Home benchmark, performed by a mobile manipulator. To make the comparison fair, the low level implementation of the skills is shared by both policies. The experiments show that with our proposed design of the FSM, the robot behaves similarly to when controlled by a BT. Since the behavior is similar, the differences between the two representations are purely in the design choice. We showed that the programming effort for such tasks is lower if BTs are preferred to FSMs, especially for large tasks that feature many behaviors and different levels of priority. 

\section*{Acknowledgments}
Authors would like to thank Michel Breyer and Julian Keller from the Autonomous Systems Lab, ETH Zürich, for the support in the implementation of the of the robot skills for the simulated robot. The experiments on the real robot were carried out in the WASP Research Arena (WARA)-Robotics, hosted by ABB Corporate Research Center in Västerås, Sweden and financially supported by the Wallenberg AI, Autonomous Systems, and Software Program (WASP) funded by the Knut and Alice Wallenberg Foundation.

\bibliographystyle{IEEEtran}
\bibliography{references}

\end{document}